\documentclass[3p,10pt]{elsarticle}
\usepackage{fullpage}
\usepackage{color}
\usepackage{amsmath,amsfonts}
\usepackage{hyperref}
\usepackage{graphicx}
\usepackage{cleveref}
\usepackage{float}
\usepackage{fancyvrb}
\makeatletter
\renewcommand{\fps@figure}{H}
\renewcommand{\fps@table}{H}
\makeatother
\usepackage{siunitx}
\usepackage{booktabs}
\usepackage[T1]{fontenc}
\usepackage{multirow}
\usepackage{tabularx}
\usepackage{lineno}
\newcolumntype{C}{>{\centering\arraybackslash}X}
\usepackage{mathtools}

\newcommand{\Vc}{\mathcal{V}}

\newcommand{\Sc}{\mathcal{S}}
\newcommand{\Tc}{\mathcal{T}}
\newcommand{\Wc}{\mathcal{W}}

\newcommand{\etab}{\boldsymbol \eta}

\newcommand{\mub}{\boldsymbol \mu}

\newcommand{\phib}{\boldsymbol \phi}

\newcommand{\cs}{\boldsymbol \sigma}

\newcommand{\1}{{\boldsymbol 1}}

\newcommand{\Ab}{{\boldsymbol A}}
\newcommand{\Bb}{{\boldsymbol B}}

\newcommand{\Fb}{{\boldsymbol F}}

\newcommand{\Nb}{{\boldsymbol N}}
\newcommand{\Pb}{{\boldsymbol P}}

\newcommand{\Tb}{{\boldsymbol T}}

\newcommand{\Vb}{{\boldsymbol V}}

\newcommand{\Xb}{{\boldsymbol X}}

\newcommand{\bb}{{\boldsymbol b}}

\newcommand{\fb}{{\boldsymbol f}}

\newcommand{\lb}{{\boldsymbol l}}

\newcommand{\nb}{{\boldsymbol n}}

\newcommand{\tb}{{\boldsymbol t}}

\newcommand{\vb}{{\boldsymbol v}}
\newcommand{\wb}{{\boldsymbol w}}
\newcommand{\xb}{{\boldsymbol x}}

\newcommand{\Rbb}{\mathbb{R}}

\newcommand{\x}{{\boldsymbol x}}

\newcommand{\X}{{\boldsymbol X}}

\newcommand{\pd}[2]{ \dfrac{\partial #1}{\partial #2}}

\usepackage{bm}

\def\pinsupport[#1](#2)(#3){
\draw (#2) + (-#3,-#3) -- +(#3,-#3) -- +(0,0) -- cycle;
\foreach \i [evaluate={\yi=(-#3+\i*(2*#3/20))},evaluate={\yj=(-#3*1.05+\i*(2*#3/20)} ] in{0,...,20}{
\draw [#1] ($(#2)-(0,#3)$) + (\yi,0) -- + (\yj,-#3/5);
};
}

\def\fixsupport[#1](#2)(#3)(#4)[#5]{
\draw [#5] (#2) + (#3,0) -- +(-#3,0) ;
\foreach \i [evaluate={\yi=(-#3+\i*(2*#3/20))},evaluate={\yj=(-#3*1.05+\i*(2*#3/20)} ] in{0,...,20}{
\draw [#1] (#2) + (\yi,0) -- + (\yj,-#4*#3/5);
};
}

\definecolor{Blue}{rgb}{0,0,1}
\definecolor{Red}{rgb}{1,0,0}
\definecolor{Orange}{rgb}{0.9,0.45,0.}
\newcommand{\KTC}[1]{}
\newcommand{\PYC}[1]{}
 \newcommand{\MMC}[1]{}
\newcommand{\rev}[1]{#1}

\newcommand{\deformationMap}{{\boldsymbol\phi}}

\newcommand{\deformationMapROM}{\tilde{\boldsymbol\phi}}
\newcommand{\deformationMapROMArgs}[3]{\deformationMapROM(#1;#2,#3)}
\newcommand{\deformationMapApprox}{{\boldsymbol g}}

\newcommand{\deformationMapApproxOne}{\tilde{\boldsymbol g}}
\newcommand{\deformationMapApproxTwo}{{\boldsymbol a}}
\newcommand{\deformationMapApproxThree}{{\boldsymbol b}}
\newcommand{\deformationMapApproxFunc}{f}

\newcommand{\deformationMapApproxArgs}[2]{\deformationMapApprox(#1;#2)}
\newcommand{\domain}{\Omega}

\newcommand{\domainROMArgs}[2]{\tilde\domain_{#1}(#2)}
\newcommand{\domainRef}{\domain_{0}}
\newcommand{\domainDef}{\domain_{t}}
 \newcommand{\params}{\boldsymbol \mu}
\newcommand{\paramDomain}{\mathcal D}
\newcommand{\paramDomainTrain}{\mathcal D_\mathrm{train}}
\newcommand{\paramDomainTest}{\mathcal D_\mathrm{test}}
\newcommand{\nparams}{q}
\newcommand{\position}{\boldsymbol x}
\newcommand{\timeArg}[1]{t_{#1}}
\newcommand{\finalTimeIndex}{T}
\newcommand{\positionArg}[2]{\position(#1,#2)}
\newcommand{\positionRef}{\boldsymbol X}
\newcommand{\positionRefDummy}{\boldsymbol X}

\newcommand{\timeDummy}{t}
\newcommand{\RR}[1]{\mathbb{R}^{#1}}
\newcommand{\RRPlus}{\mathbb{R}_{+}}
\newcommand{\genCoordsDummy}{\hat{\boldsymbol y}}
\newcommand{\genCoords}{\hat{\boldsymbol x}}

\newcommand{\genCoordsArgs}[2]{\genCoords(#1,#2)}
\newcommand{\genCoordsArgsParam}[2]{\genCoords_{\theta_e}(#1,#2)}

\newcommand{\genVels}{\hat{\boldsymbol v}}
\newcommand{\nred}{r}

\newcommand{\manifold}{\mathcal M}

\newcommand{\defeq}{:=}
\newcommand{\deformationGradient}{\boldsymbol F}
\newcommand{\deformationGradientApprox}{\tilde{\deformationGradient}}

\newcommand{\timeInterval}{\mathcal T}

\newcommand{\vg}{\boldsymbol{g}}
\newcommand{\enc}{\boldsymbol{e}}
\newcommand{\vx}{\boldsymbol{x}}
\newcommand{\vX}{\boldsymbol{X}}

\newcommand{\phibDummy}{\boldsymbol \psi}

\newcommand{\nParticles}{P}
\newcommand{\nparticles}{\nParticles}
\newcommand{\admissibleTest}{\Vc_0} %

\newcommand{\nBasis}{{B}}
\newcommand{\nTimesteps}{{T}}
\newcommand{\sumParticles}{\sum_{p=1}^{\nParticles}}
\newcommand{\sumQuads}{\sum_{p=1}^{\nQuad}}
\newcommand{\sumBasis}{\sum_{i=1}^\nBasis}
\newcommand{\sumjBasis}{\sum_{j=1}^\nBasis}
\newcommand{\timestepn}{\Delta t_n}
\newcommand{\manifoldX}{\manifold(\X)}
\newcommand{\sampleSet}{\mathcal P}
\newcommand{\neighborSet}{\mathcal N}
\newcommand{\quadPointSet}{\{1,\ldots,\nQuad\}}
\newcommand{\inquadPointSet}{=1,\ldots,\nQuad}
\newcommand{\basisFunctionSet}{\mathcal I}
\newcommand{\sumBasisFunctionSet}{\sum_{i\in\basisFunctionSet}}

\newcommand{\sumParticlesSampleOnly}{\sum_{p\in\sampleSet}}

\newcommand{\radcyl}{r_\text{c}}
\newcommand{\radpok}{r_\text{p}}

\usepackage[ruled,vlined,linesnumbered]{algorithm2e}
\SetKwInput{KwData}{Input}
\SetKwInput{KwResult}{Output}

\usepackage{amsthm}

\newtheorem*{remark}{Remark}

\newcommand{\cw}{\Delta x}

\newcommand{\mass}{m^p}
\newcommand{\massPi}{m^{\quadmap(p)}}
\newcommand{\velTn}{\vb^p_n}
\newcommand{\velTnPi}{\vb^{\quadmap(p)}_n}
\newcommand{\velTnp}{\vb^p_{n+1}}
\newcommand{\velTnpTrial}{\vb^{p,\text{trial}}_{n+1}}
\newcommand{\xTnpTrial}{\xb^{p,\text{trial}}_{n+1}}
\newcommand{\pointTn}{\xb^p_n}
\newcommand{\quadmap}{\Pi}
\newcommand{\pointTnPi}{\xb^{\quadmap(p)}_n}
\newcommand{\FTn}{\Fb^p_n}
\newcommand{\FTnPi}{\Fb^{\quadmap(p)}_n}
\newcommand{\FTnp}{\Fb^p_{n+1}}

\newcommand{\quadMass}{m^{Q,p}}
\newcommand{\quadVol}{V^{Q,p}}
\newcommand{\cellVol}{V^c}
\newcommand{\quadMassTn}{m^{Q,p}_n}

\newcommand{\quadPoint}{\vx^{Q,p}}
\newcommand{\quadPointIni}{\vX^{Q,p}}
\newcommand{\quadVelTn}{\vb^{Q,p}_n}
\newcommand{\quadPointTn}{{\vx^{Q,p}_n}}
\newcommand{\quadPointIniTn}{{\vX^{Q,p}_n}}
\newcommand{\quadFTn}{{\Fb^{Q,p}_n}}
\newcommand{\quadJTn}{{J^{Q,p}_n}}

\newcommand{\nQuad}{{P^{Q}}}

\newcommand{\myparagraph}[1]{\paragraph{#1}\mbox{}}

\def\appendixname{}
\biboptions{numbers,sort&compress}

\usepackage{todonotes}

\begin{document}

\title{Model reduction for the material point method via an \\ implicit neural
representation of the deformation map}

\author[columbia]{Peter Yichen Chen\corref{colcorr}}
\cortext[colcorr]{Corresponding author}
\ead[url]{peterchencyc.com}
\ead{cyc@cs.columbia.edu}
\author[facebook]{Maurizio M. Chiaramonte}
\ead{mchiaram@fb.com}
\author[toronto,columbia]{Eitan Grinspun}
\ead{eitan@cs.toronto.edu}
\author[facebook]{Kevin Carlberg}
\ead{carlberg@fb.com}

\address[columbia]{Columbia University, 116\textsuperscript{th} St $\&$ Broadway, New York, NY 10027, USA}
\address[facebook]{Meta Reality Labs Research, 
9845 Willows Road, Redmond, WA 98052, USA}
\address[toronto]{University of Toronto, 40 St. George Street, Room 4283, Toronto, ON M5S 2E4, Canada}

\begin{abstract} 
	This work proposes a model-reduction approach for the material point method on nonlinear manifolds. Our technique approximates the $\textit{kinematics}$ by approximating the deformation map using an implicit neural representation that restricts deformation trajectories to reside on a low-dimensional manifold. By explicitly approximating the deformation map, its spatiotemporal gradients---in particular the deformation gradient and the velocity---can be computed via analytical differentiation. In contrast to typical model-reduction techniques that construct a linear or nonlinear manifold to approximate the (finite number of) degrees of freedom characterizing a given spatial discretization, the use of an implicit neural representation enables the proposed method to approximate the $\textit{continuous}$ deformation map. This allows the kinematic approximation to remain agnostic to the discretization. Consequently, the technique supports dynamic discretizations---including resolution changes---during the course of the online reduced-order-model simulation.
	
	To generate $\textit{dynamics}$ for the generalized coordinates, we propose a family of projection techniques. At each time step, these techniques: (1) Calculate full-space kinematics at quadrature points, (2) Calculate the full-space dynamics for a subset of `sample' material points, and (3) Calculate the reduced-space dynamics by projecting the updated full-space position and velocity onto the low-dimensional manifold and tangent space, respectively. We achieve significant computational speedup via hyper-reduction that ensures all three steps execute on only a small subset of the problem's spatial domain. Large-scale numerical examples with millions of material points illustrate the method's ability to gain an order of magnitude computational-cost saving---indeed $\textit{real-time simulations}$---with negligible errors.

\end{abstract}

\maketitle

\section*{Highlights}
\begin{itemize}
   \item Novel model-reduction technique for the material point method
	 \item \textit{Kinematics}: implicit neural representation of the deformation map
	 \item \textit{Dynamics}: projection of position and velocity onto manifold and
		 tangent space, respectively
	 \item \textit{Hyper-reduction}: achieve computational-cost savings by
		 computing full-space dynamics for a small number of material points
		 \KTC{Note that this was previously `which requires tracking a small
		 number of material points', which isn't strictly true for Eulerian
		 quadrature.}
	 \item Deformation-map approximation supports super-resolution and adaptive
		 Eulerian quadratures
   \item Numerical experiments demonstrating an order of magnitude speedup
\end{itemize}

\section{Introduction}
Computational physics plays a pivotal role in modern-day science and
engineering, with important applications spanning physics, chemistry, material
science, civil engineering, aerospace engineering, visual effects, virtual reality, gaming, and many more. In these domains, practitioners must address the
\textit{fidelity--cost tradeoff}. In particular, to ensure computational
models satisfy the verification and validation standards intrinsic to the
application at hand, practitioners must generate high-fidelity models
characterized by a sufficiently fine spatiotemporal resolution. In many
cases---especially for high-consequence applications with stringent
requirements on predictive fidelity---this leads to highly resolved models whose computational cost
precludes them from being employed in time-critical applications such as
real-time data assimilation, fast-turnaround design under uncertainty, and
interactive simulations. Such applications demand rapid simulation times, with
\textit{real-time simulation} a requirement in some cases. This leads to a
\textit{computational barrier}: sufficiently accurate computational models are
often too computationally costly to be deployed in important time-critical
applications, which necessitates the use of simplified models in such cases, which---in turn---often violate the accuracy requirements of the
application. 

In this work, we propose to overcome this computational barrier for a widely
adopted simulation framework in continuum mechanics: the material point method
(MPM). To achieve this, we propose a novel projetion-based model-reduction
method that leverages implicit neural representations of the deformation map.
To our knowledge, this work comprises the first time a model-reduction
technique has been proposed for MPM or any other point-cloud-based simulation
techniques, e.g., smoothed-particle hydrodynamics (SPH). We proceed by
reviewing the literature for projection-based model reduction and the material point method in Sections
\ref{sec:romreview} and \ref{sec:mpmreview}, respectively, followed by a
summary of our contributions in Section \ref{overview}.
\KTC{I reshuffled the intro to have a more logical flow}

\subsection{Projection-based model reduction}\label{sec:romreview}
To address the computational barrier mentioned above for a range of computational
methods, researchers have pursued projection-based model-reduction techniques
\citep{benner2015survey}. In contrast to more common approaches to model
simplification (e.g., coarse graining, linearization), such techniques attempt
to inherit the benefits of high-fidelity models (e.g., fine resolution,
rich constitutive laws, material/geometric nonlinearities, dynamical-system properties such
as symplecticity) while drastically reducing simulation costs by restricting
trajectories to evolve on a low-dimensional subspace or manifold. When applied
successfully, these reduced-order models can incur orders-of-magnitude savings
in computational cost while incurring negligible errors.  Reduced-order models have been successfully employed to
solve real-world problems in many fields, such as motor-generator design
\citep{bruns2015parametric}, batch chromatography \citep{benner2015reduced},
fluid dynamics \citep{hall2000proper,
willcox2002balanced,bergmann2005optimal,lieu2006reduced,carlberg2013gnat,mainini2015surrogate,carlbergGalDiscOpt},
structural dynamics \citep{amsallem2009method}, computer graphics
\citep{james2006precomputed,barbivc2011real,yang2015expediting}, and robotics
\citep{tan2020realtime}.

Model reduction for dynamical systems dates back to
\citet{sirovich1987turbulence}, who applied principal component analysis (PCA)
to turbulence simulations and coined the term proper orthogonal decomposition
(POD). Model-reduction methods typically require two stages: an \emph{offline}
or `training' stage, and an \emph{online} or `evaluation' stage. The
offline stage executes costly computations in order to generate a
low-dimensional subspace or manifold to approximate the system's
kinematics; in the case of POD, this
corresponds to executing many expensive high-fidelity simulations at different
problem-parameter instances, computing the singular value decomposition of
resulting solution snapshots, and preserving the dominant left singular
vectors as a basis for a low-dimensional subspace. The online stage
executes rapid simulations by projecting the system's dynamics onto the
low-dimensional subspace or manifold in a manner that preserves key
dynamical-system properties; if the dynamical-system operators are nonlinear,
then hyper-reduction techniques are employed to ensure computational-cost
savings, in which only a subset of the problem domain is employed to perform
projection \citep{ryckelynck2005priori, nguyen2008efficient,
galbally2010non,amsallem2012nonlinear,drohmann2012reduced,carlberg2013gnat}.

Traditionally, model-reduction techniques have employed linear subspaces for
kinematic approximation; this includes the
aforementioned POD method \citep{sirovich1987turbulence,
everson1995karhunen,lall2002subspace, rathinam2003new, SIRISUP200492,
rowley2005model, barone2009stable, bergmann2009enablers,carlberg2011low,
wang2012proper,holmes2012turbulence,peherstorfer2015online,carlberg2015adaptive,
carlberg2015preserving,abgrall2018model}, the reduced-basis technique
\citep{Prud'homme200270, Rozza20071}, balanced truncation \citep{Moore198117},
rational interpolation \citep{baur2011interpolatory, gugercin2008h_2},
Craig--Bampton model reduction \citep{craig1968coupling}, and least-squares
Petrov--Galerkin projection
\citep{carlberg2011efficient,FANG2013540,carlberg2013gnat,carlbergGalDiscOpt}. Recently, model
reduction on nonlinear manifolds has gained increased attention
\citep{8062736,7799153,gu2011model,erichson2019physics,maulik2020time,regazzoni2019machine,fulton2019latent,kim2019deep,lee2020model,lee2019deep,maulik2021reduced,romero2021learning,shen2021high}.
In particular, for problems characterized by a slowly decaying Kolmogorov width
(e.g., advection-dominated problems), nonlinear manifolds---often constructed
with deep neural networks---have been shown to outperform their linear
counterparts significantly. This is due to two factors: the theoretical
ability of nonlinear manifolds to overcome the Kolmogorov-width limitations of
linear subspaces, and the recent development of deep-learning tools
\cite{paszke2019pytorch} that facilitate generating accurate nonlinear
manifolds with requisite smoothness properties from data \cite{lee2020model}.

Relatedly, data-driven dynamics-learning methods can also be used to generate
approximations of high-fidelity computational models as an alternative to
projection-based reduced-order models
\cite{lusch2018deep,morton2018deep,otto2019linearly,takeishi2017learning}.
These techniques aim to learn both the embedding (i.e., mapping from
high-dimensional state to low-dimensional latent variables) and the dynamics
(i.e., the time evolution of these latent variables) in a purely data-driven
manner that requires only observing the state and/or velocity during training
simulations. \rev{Similarly, data-driven surrogate models \citep{umetani2014pteromys,martin2015omniad,dosovitskiy2015learning,schulz2017interactive,chen2019visual} train direct mappings from problem parameters to physical states. Both data-driven dynamics-learning methods and surrogate models are extremely fast to compute due to their data-driven nature since no solver is required at inference time.} However, since these techniques do not require explicit knowledge of
the equations governing the dynamics of the system, they suffer from a range of drawbacks, including violation of
important physical properties underpinning the dynamical system, challenges in
performing error analysis and control, and a lack of generalization and
robustness. For these reasons, the current work focuses on projection-based
model reduction.

\subsection{Material point method}\label{sec:mpmreview}
MPM was introduced by \citet{sulsky1995application} as an
extension of the particle-in-cell (PIC) method for solid mechanics; it is a
hybrid Eulerian--Lagrangian discretization method widely employed in solid,
fluid, and multiphase simulations. Due to its dual Eulerian and Lagrangian
representations, MPM offers several advantages over the finite element method
(FEM), such as its ability to more easily handle problems characterized by
large deformation, fracture, contact, and collisions
\cite{bardenhagen1998shear,
wikeckowski1999particle,nairn2003material,patankar2001lagrangian,york2000fluid,li2002meshfree,stomakhin2013material,jiang2017anisotropic,ram2015material,klar2016drucker,gast2015optimization,stomakhin2014augmented,jiang2015affine,yue2015continuum,yue2018hybrid,
chen2021hybrid,daviet2016semi,mast2014avalanche,daphalapurkar2007simulation,sadeghirad2011convected,fang2019silly,wolper2019cd,han2019hybrid,wang2019simulation,fang2020iq,jiang2020hybrid,
li2020soft, fei2021revisiting,su2021unified}.

However, MPM's dual Eulerian--Lagrangian representation of the material and
the requisite transfer between these representations also make it very
computationally costly.\KTC{Can you provide a more complete description for
why this makes it costly? E.g., the transfer, tracking particles, need for
many particles, etc.}\PYC{done.} In particular, MPM typically tracks a large number of
Lagrangian material points, which can be loosely interpreted as particles. At every time step, to compute the dynamics update of
these material points, MPM transfers the particle information onto the Eulerian grid
and conducts the dynamics update on the grid. Subsequently, MPM transfers the
updated velocities back to the Lagrangian particles. Consequently, MPM's
computational cost is larger than either a strictly Lagrangian approach or a strictly
Eulerian approach. Recent advances in sparse data structures
\citep{gao2018sparse,hu2019hybrid}, compiler optimization
\citep{hu2018moving,hu2019taichi,hu2021quantaichi}, and multi-GPU
\citep{gao2018gpu,wang2020massively} have made substantial progress in
alleviating the computational cost of MPM, leading to practical applications
of MPM to areas such as robotic control
\citep{hu2019chainqueen,hu2019difftaichi} and topology optimization
\citep{li2021lagrangian}. Yet, real-time, million-particle MPM simulations
remain out of reach.\KTC{I rewrote the previous sentence, please check you
agree.}\PYC{edited.} We aim to address this computational barrier by
developing a novel model-reduction method tailored to MPM.

Prior work on model reduction techniques for MPM is scarce if it exists at
all; literature on model reduction for alternative flavors of PIC methods and other point-cloud-based simulation techniques is also
severely limited. The few exceptions include the following contributions:
\citet{nicolini2019model} applied POD to the PIC-based solver of the
Maxwell--Vlasov equations, and \citet{wiewel2019latent} used convolutional
neural networks (CNNs) to reduce the dimension of the Eulerian grid data of
the fluid implicit particle (FLIP) method and used a long short-term memory
(LSTM) networks to evolve the subspace.\KTC{Peter, can you quickly summarize
why these are insufficient or why we want to improve beyond them?}\PYC{done.}
However, these works focus on fluid mechanics problems and only conduct model
reduction for the Eulerian degrees of freedom. By contrast, MPM is
particularly designed for solid mechanics, and model reduction for the
Lagrangian degrees of freedom has to be addressed. Relatedly, graph neural network (GNN) has also been used to model physical systems with MPM training data \citep{sanchez2020learning}. However, since GNN reduces neither the dimensionality of the system nor the
complexity of the simulation, it offers no
computational-cost advantages over the original high-fidelity MPM simulations.

\KTC{I recommend deleting this paragraph or just extracting the part about MPM
approximations and tying it in above. Thoughts, Peter?}\PYC{agree! done.} \KTC{Peter: I haven't read these works; why are they
used in the first place if they aren't faster?}\PYC{just to show machine learning has the ability to learn complex things. these works are very popular so it is worth metnioning them even if they are slower. Youngsoo onced invited the authors to DDPS seminar.}

\subsection{Overview of contribution}
\label{overview}
To develop a model-reduction framework for MPM, we first notice that MPM is
characterized by \textit{discrete Lagrangian kinematics}, as kinematic
information is stored on material points, and \textit{Eulerian dynamics}, as
force calculations are performed on a background Eulerian grid. As such, we
must develop a model-reduction framework that is compatible with this
conceptual decomposition.

\begin{figure}[H]
    \centering
    \includegraphics[width=\textwidth]{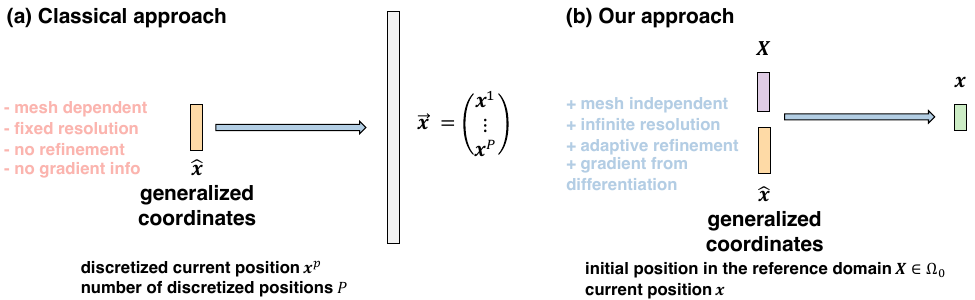}
    \caption{Our approach vs. the classical approach. (a) In classical model
		reduction techniques, a mapping from the generalized coordinates
		$\genCoords$ is often trained to infer the deformed positions $\vx^p$ of a
		finite number $\nParticles$ of particles concatenated into a column
		vector. Since this low-dimensional subspace is constructed for the
		discretized positions of the original continuous deformation map, it has
		several key limitations: (1) it is mesh-dependent; (2) it does not support
		resolution change; (3) it cannot handle adaptive resolutions during
		simulation; (4) it does not provide gradient information about the
		deformed positions. (b) By contrast, our approach builds an implicit
		neural representation of the deformation map. More precisely, this
		representation
		corresponds to a manifold-parameterization function that maps $\genCoords$
		and an arbitrary undeformed position $\vX$ to its deformed position $\vx$.
		Consequently, we can represent an infinite number of particle positions,
		i.e., the entire deformation map, using the finite-dimensional generalized
		coordinates $\genCoords$. In other words, we built a low-dimensional
		approximation of the continuous deformation map itself instead of the
		discretization of the deformation map as it is done in the classical
		approach. Consequently, we can address all four aforementioned
		limitations. \KTC{On the left, change `no resolution change' to `fixed
		resolution'; `inadaptive simulation' to `no refinement possible'.}\PYC{done.}}
    \label{img:vs_classic}
  \end{figure}

To achieve this, we perform \textit{Lagrangian kinematics approximation}. In principle, we could achieve this in the canonical
way by constructing a (linear or nonlinear) mapping from low-dimensional
generalized
coordinates (i.e., latent variables) to the position of all material
points in a high-fidelity MPM discretization as depicted in
\Cref{img:vs_classic}a. However, this introduces two major challenges. First,
computing the deformation gradient required for stress calculations becomes
challenging; the deformation gradient would need to be computed on the
Eulerian grid and then transferred to the tracked material points, which could
lead to inconsistencies between the advected deformation gradient and the
kinematic approximation itself. Second, hyper-reduction would become very
difficult, as all neighboring material points that could \textit{ever} influence the
(Eulerian) dynamics of the tracked set of material points over the entire
trajectory would need to be identified \textit{a priori}; this is not possible
to do for general trajectories.  In addition to these challenges, the kinematic
approximation is `tied' to a specific, pre-defined spatial discretization,
precluding dynamic resolution changes that might be advantageous to
introduce during the reduced-order-model simulation.

Thus, we develop a novel kinematic approximation that directly approximates the
continuous deformation map itself; architecturally, this implies that the
input to the parameterization function includes \textit{both} the generalized
coordinates and the reference-domain coordinates of the material point of
interest, with the output corresponding to the deformation of that material
point under the configuration imposed by the generalized coordinates.
\Cref{img:vs_classic}b depicts this kinematic approximation, which is
tantamount to constructing an implicit neural representation of
the deformation map.  The resulting
approximation is independent of the high-fidelity discretization by construction, as it effectively learns a
mapping between the generalized coordinates and the deformation that is
applicable to material points associated with
\textit{any point} in the reference domain. Consequently, the kinematics of
arbitrary material points, including the deformation gradients and the velocities, can be recovered from the approximation. This
mesh-independence feature enables dynamic resolution during the ROM
simulation, even super-resolution, wherein additional material points that
were not
present during training can be introduced online.

		In essence, our low-dimensional manifold is an implicit neural
		representation of the deformation map. Implicit neural representation, a
		robust representation of arbitrary vector fields, has found substantial
		recent success in the computer-vision community, and has been shown to 
		generate accurate approximations of signed distanced fields
		\citep{park2019deepsdf,mescheder2019occupancy,chen2019learning}, image
		channels \citep{sitzmann2020implicit}, as well as radiance fields
		\citep{mildenhall2020nerf}. Thanks to its continuous nature, implicit
		neural representation has infinite resolution and continuous
		differentiability, both of which are crucial for solving the dynamics of
		physical systems, where adaptive quadratures and gradient computation are
		frequently required. 

		To our knowledge, the concurrent work by \citet{pan2022neural} and our work are the first time implicit neural representations have been used for model reduction for any physical system. Alternatively, our low-dimensional approximation can also be viewed as an
		extension of the physics-informed neural network
		(PINN) \citep{raissi2019physics} for model reduction. PINN explicitly
		models time using a one-dimensional variable $t$. By contrast, our
		approach models time implicitly via high-dimensional generalized
		coordinates $\genCoords(t)$. Indeed, when $\genCoords(t)=t$, our model
		recovers the exact formulation of PINN. By implicitly modeling time $t$, our
		representation enables online simulations that undergo drastically
		different temporal trajectories than the original offline training
		simulations. Such a feature is crucial for applications involving diverse
		user interactions with the physical system.

		After the low-dimensional manifold is constructed, we perform
		\textit{Eulerian dynamics approximation}.  Specifically, at each time
		step, the method (1) calculates full-space kinematics at quadrature
		points, (2) calculates the full-space dynamics by computing position and
		velocity updates in the full space for a subset of `sample' material
		points, and (3) calculates the reduced-space dynamics by projecting the
		updated full-space position and velocity onto the low-dimensional manifold
		and tangent space, respectively. In the first step, we consider both
		Lagrangian and (adaptive) Eulerian quadrature rules. The latter of these
		is enabled by the invertibility of the deformation-map approximation and
		facilitates hyper-reduction, as it obviates the need to identify and track
		neighboring material points that influence the dynamics of the `sample'
		material points.

    The remainder of the paper is organized as follows. First, 
    \Cref{sec:full} summarizes the fundamentals of the material point method.
    Next, \Cref{sec:rom} introduces the proposed model-reduction approach,
    including the kinematic approximation (\Cref{sec:mpkinematics}), the
    dynamics approximation (\Cref{sec:dynamics}), and the approach to
    hyper-reduction (\Cref{sec:hyper-reduction}). Then, \Cref{sec:train}
    describes the practical design of the kinematic approximation, including the
    architecture choice for the associated neural network. \Cref{sec:experiments} reports numerical experiments,
    and---finally---\Cref{sec:conclusions} concludes the paper.

\section{Full-order model}
\label{sec:full}
As the material point method\KTC{I think we should avoid capitalizing MPM
throughout} is a hybrid Eulerian--Lagrangian method, we will introduce both formulations of the problem statement. This section first introduces the full-order continuous problem statement 
in Section~\ref{subsxn:problem_statement} and later discretizes it using MPM in Section~\ref{subsxn:mpmformulation}.

\subsection{Continuous problem formulation} 
\label{subsxn:problem_statement}

We study the trajectory of a solid body with a reference configuration 
given by $\domainRef$ during the time interval
$\timeInterval\defeq[\timeArg{0},\timeArg{\finalTimeIndex}]\subseteq\RR{}$
such that the body at any time $t$ occupies a domain $\domainDef$. In what
follows we use $\partial \Omega$ to denote the boundary of the domain
$\Omega$\KTC{note slight abuse of notation using both $\Omega$ and $\Omega_t$}. We decompose the boundary as $\partial
\Omega = \overline{\partial_N \Omega \cup \partial_D \Omega}$, $\partial_N
\Omega \cap \partial_D \Omega = \emptyset$ with $\partial_N \Omega $ and  $\partial_D \Omega $ denoting the portions of the
boundary with prescribed Neumann and Dirichlet boundary conditions, respectively.

We restrict attention to hyperelastic materials such that there exists
a potential function of the deformation gradient from which we can derive
internal stresses \citep{holzapfel2002nonlinear}. Additionally, we assume that problem parameters (e.g.,
geometric parameters, boundary conditions, external forces) can be represented
by the parameter vector $\mub \in \paramDomain$, where
$\paramDomain\subseteq\RR{\nparams}$ denotes the parameter domain. In the
remainder of Section \ref{sec:full}, 
we omit the explicit dependency on $\mub$ for simplicity of
exposition; we reintroduce parameter dependence in Section \ref{sec:rom} to
emphasize the parameterized evaluation of the proposed reduced-order model.

\subsubsection{Lagrangian strong form}
\label{subsubsxn:lagrangian_sf}
We define the deformation map $\deformationMap:\domainRef\times
\timeInterval\rightarrow \domainDef$, as the mapping from any
point on the
undeformed domain\KTC{reference domain?}\PYC{interchangeable} $\domainRef$ to the
corresponding point on the deformed domain $\domainDef$ at a time $t\in\timeInterval$.
To enforce initial conditions and essential (Dirichlet) boundary conditions, we restrict the deformation map $\phib$ to
reside in the space of admissible trajectories $\Sc$ such that $\phib \in \Sc$, where\footnote{Note that we can prescribe either displacements or velocities as essential boundary conditions. We restrict boundary displacements to be smooth in time. Therefore, the boundary displacements are uniquely defined by prescribed boundary velocities. Thus in our formulation, without loss of generality, we can consider velocity-only essential boundary conditions.}
\begin{align}
\begin{split}
    \Sc \defeq \{ \phibDummy:\domainRef\times \timeInterval\rightarrow \RR{d}\, |\, &
    \phibDummy( \positionRef, 0 ) = \positionRef\ \text{  and  } \
    \dot \phibDummy(\positionRef,0) = \overline \Vb(\positionRef, 0),\ \forall \Xb\in\domainRef ; \\
    &\dot \phibDummy(\positionRef,t) =
    \overline \Vb(\positionRef, t), \, \forall \positionRef \in \partial_D
    \domainRef(t), \ \forall t\in\timeInterval
    \}. 
\end{split}
\end{align}
Here, $\dot{()}$ denotes differentiation with respect to time for a fixed
position on the reference domain $\positionRef\in\domainRef$, also known as the 
material time derivative; $\overline \Vb$ 
denotes both the prescribed initial velocity $\overline \Vb(\cdot, 0)
:\domainRef \to \Rbb^d
$
and the prescribed boundary velocities 
$\overline \Vb
:\partial_D\Omega_0(t) \times \timeInterval \to \Rbb^d $. \KTC{MMC: tried to
clean this up a bit, thoughts?}

The problem then becomes: Find $\deformationMap \in \Sc$ such that for all time $t\in\timeInterval$ 
\begin{align} 
    \rho_0 \ddot \deformationMap = \nabla_{\X} \cdot \Pb(\nabla_\X \deformationMap )  + \Bb, \quad & \forall \X \in \domainRef, \label{eq:strong_form_lag} \\
    \Pb\Nb = \overline \Tb,  \quad & \forall \X \in \partial_N \domainRef,
\end{align} 
where $\rho_0$ is the initial density defined on the reference domain, $\Pb$
denotes the first Piola--\rev{Kirchhoff} stress tensor, $\overline \Tb$ and $\Bb$ are
external tractions and body forces, respectively, and $\Nb$ denotes the normal
to the boundary $\partial \domainRef$.  \KTC{MMC: should we have 
$\forall \X \in \partial_N \domainRef(t)$ in the second equation above to
emphasize that the Dirichlet boundary may change in time?}

\subsubsection{Eulerian strong form}

We can reformulate the problem of Section \ref{subsubsxn:lagrangian_sf} in an
Eulerian (i.e., spatially fixed) reference frame.
Note that for our particular problem, 
we need to retain a notion of deformation between adjacent material points as stress depends on the deformation gradient. Hence in what follows, we formulate the 
traditional Cauchy's equations of motion, with velocity being the primary unknown variable, augmented by an advection 
equation to ``transport'' deformation gradient along with the flow of the body.

The primary unknowns become the spatial velocity $\vb$ and deformation gradient $\Fb$ that belong to their respective admissible sets 
\begin{align}\label{eq:eulerian_admissible}
    \Vc &= \{ \wb : \domainDef \times \timeInterval \to \Rbb^d \ |\  \wb(\x, 0
		) = \overline \vb(\x, 0), \; \forall \x \in \domainRef; \wb(\x, t) =
		\overline \vb(\x, t ), \; \forall (\x, t)  \in \partial_D \domainDef \times \timeInterval \},  \\ 
    \Wc &= \{ \Ab: \domainDef \times \timeInterval \to \Rbb^{d \times d }\  |\
		\Ab(\X, 0 ) = \1, \; \forall \X \in \domainRef \},
\end{align}
where $\1$ denotes a diagonal matrix of ones, and $\overline \vb( \phib(\X,
t), t) = \overline \Vb(\X, t)$ defines both initial and boundary conditions. 

The strong formulation of the problem statement, in the Eulerian frame of reference, can then be expressed as follows: Find the velocity $\vb \in \Vc$ and the deformation gradient $\Fb \in \Wc$ 
such that for all $t\in\timeInterval$
\begin{align}
    \rho \dot \vb = \nabla_{\x} \cdot \cs(\Fb)  + \bb, \quad & \forall \x \in \domainDef, \label{eq:strong_form_eul} \\
    \cs\nb = \overline \tb,  \quad & \forall \x \in \partial_N \domainDef,
\end{align} 
and
\begin{align} 
\dot \Fb = \pd{\Fb}{t} + (\nabla_\x \Fb) \vb = (\nabla_{\x} \vb) \Fb, \quad
	\forall \x \in \domainDef,
\label{eq:advection_def_grad}
\end{align}
where $\cs$ denotes the Cauchy stress tensor related to the first Piola--\rev{Kirchhoff} tensor by $\Pb = J \cs \Fb^{-\top} $ with $J = \det(\Fb)$, $\bb:\domainDef\rightarrow \RR{d}$ denotes body forces, 
  $\overline
\tb:\partial_N \domainDef\rightarrow\RR{d} $ denotes the prescribed tractions, and $\rho:\domainDef\rightarrow\RR{}_+$ denotes the material density
\footnote{Body forces, tractions, and densities can be related to their corresponding Lagrangian
 quantities as follows $J \bb = \Bb$, $\overline \tb \| J \Fb^{-\top} \Nb \| = \overline \Tb$,
  $ J \rho =  \rho_0 $.}. The mapping $\phib$ can be recovered by integrating in time

\begin{align} 
\phib(\X, t)   =  \X + \int_{0}^{t} \vb( \x , \tau)  d\tau .
\label{eq:def_map_int}
\end{align} 

\PYC{I like the overall perspective where we have a solid continuous understanding of both the lagrangian and the eulerian formulations. Previous MPM discussions on the eulerian formulation are handwavy, e.g., missing boundary conditions, deformation gradient, relationship with deformation map, etc. So this is good!}
\KTC{Agreed, nice work Maurizio! Going above and beyond as usual!}

\subsection{MPM discretization}
\label{subsxn:mpmformulation}
We begin by discretizing the time interval $\Tc$ in discrete time instances $\{ t_n \}_{n=0}^{\nTimesteps}$, where a subscript $n$ denotes a quantity defined at time step $n$. In the following sections, we first present how we approximate the solution of Eq.~\eqref{eq:def_map_int} 
and Eq.~\eqref{eq:advection_def_grad}. Next, we describe the discretization of
the Eulerian equations of motion Eq.~\eqref{eq:strong_form_eul}.

\subsubsection{Lagrangian discretization} 

We discretize our domain $\Omega_0$ with a collection of particles of finite volumes and masses 
$\{ \X^p \}_{p=1}^{\nParticles}$  which at any time $t_n$ occupy positions $\{ \x^p_n \}_{p=1}^{\nParticles}$ and have 
mass $\{ m^p \}_{p=1}^{\nParticles}$. In practice this can be achieved, for example, by generating a simplicial 
subdivision of $\Omega_0$, assigning $\X^p \equiv \x^p_0$ as the barycenter of
the $p$th simplex,
 and $m^p$ 
the volume of the $p^\text{th}$ simplex times the density $\rho_0(\X^p)$.

At each time step, given $\vb(\xb, t_n)$, we can 
evaluate $\vb_n^p := \vb(\xb^p_n, t_n)$ as well as $\lb_n^p := \nabla_{\x} \vb(\xb_n^p, t_n)$. With the above 
we can integrate in time Eq.~\eqref{eq:def_map_int} and Eq.~\eqref{eq:advection_def_grad} to obtain
\begin{align}
\xb_{n+1}^p &= \xb^{p}_{n} + \Delta t_n \vb^{p}_n \\
\Fb^{p}_{n+1} &= \Fb^p_n + \Delta t_n \lb_{n}^p \Fb^{p}_n
	,
\end{align}
for $n=0,\ldots,\nTimesteps-1$, where $\Delta t_n \defeq t_{n+1} - t_n$. 

\subsubsection{Eulerian discretization}

Assuming sufficient regularity, an equivalent weak formulation of the Eulerian
strong form Eq.~\ref{eq:strong_form_eul} is: Find the velocity $\vb \in \Vc$
such that for all time $t\in\timeInterval$ 
\begin{align}\label{eq:weakFormTwo}
\begin{split}
    &\int_{\Omega} \rho \dot \vb \cdot \etab dV = \int_{\Omega}  (\bb \cdot \etab - \cs(\Fb):\nabla
    \etab ) dV + \int_{\partial_N
    \Omega} \overline \tb \cdot \etab ds  \quad \forall \etab \in\admissibleTest, \\
\end{split}
\end{align}
where $\admissibleTest$ denotes the set of admissible test functions at time
$t$, defined as 
\[
    \admissibleTest \defeq  \{ \wb : \domainDef \times \timeInterval \to
		\Rbb^d \, |\,  \wb(\x, 0 ) = 0, \; \forall \x \in \domainRef; \wb(\x, t) =
		0, \; \forall (\x, t)  \in \partial_D \domainDef \times \timeInterval \}.
\]
The above can be recast in mass-integral form using the relation $dm = \rho dV$ as
\begin{align}
\begin{split}
    &\int_{\Omega} \dot \vb \cdot \etab dm =  \int_{\Omega} \frac{J}{\rho_0}  (\bb \cdot \etab - \cs(\Fb):\nabla \etab ) dm + \int_{\partial_N
\Omega} \overline \tb \cdot \etab ds, \quad \forall \etab \in\admissibleTest , \; \forall t \in \Tc, 
\end{split} \label{eq:massWeakForm}
\end{align}
where $J := \det(\Fb) $. %
We further assume a finite-dimensional approximation of $\Vc$ by 
\[
    \Vc^h = \{ \wb^h \in \Vc\, |\, \wb^h = \sumjBasis \wb(t) N_j(\x) \},
\]
and a similar finite-dimensional approximation for $\Vc_0$. We therefore can
express the entire velocity field via a finite number of (Eulerian) basis functions $\vb(\x, t) \approx \sumjBasis \vb_{j} N_j(\x) \in \Vc^h$. Combining this with the relation 
\begin{equation}
    \label{eqn:material_point}
\int_{\Omega}( \bullet) dm \approx \sumParticles  (\bullet) \mass,
\end{equation}
we arrive at the set of discrete equations 
\begin{equation} 
\sumParticles (  \sumjBasis \dot \vb_j N_j \, N_i )|_{\x^p} \;  \mass = 
   \sumParticles \frac{1}{\rho_0}\left[ J \; (  \bb N_i - \cs( \Fb ) \nabla N_i )
   \right]|_{\x^p} \mass + \int_{\partial_N
   \Omega} \overline \tb N_i, \quad
   i=1,\ldots,\nBasis
   \label{eq:weak_form_discrete_space} .
\end{equation}

By invoking the mass-lumping approximation, we approximate the left-hand side of
Eq.~\eqref{eq:weak_form_discrete_space} as
\begin{equation}\label{eq:spaceDiscTwo}
\sumParticles (  \sumjBasis \dot \vb_j N_j \, N_i )|_{\x^p} \;  \mass
= \sumParticles (  \sumjBasis  N_j \, N_i )|_{\x^p} \;  \mass \dot \vb_j =
\sumjBasis M_{ij} \dot \vb_j \approx m_i \dot \vb_i,\quad i=1,\ldots,
   \nBasis,
\end{equation}
where
$M_{ij}\defeq (\sumParticles N_jN_i)_{\x^p}\mass$ and $m_i\defeq\sumjBasis M_{ij}$.

Combining the spatial discretization above with time discretization, we now have enough ingredients to devise
an explicit time-integration scheme; Algorithm \ref{alg:MPM} reports the resulting
algorithm that employs the symplectic Euler method.

\begin{algorithm}[htb!]
\SetAlgoLined
\KwData{Deformation gradient $\FTn$, velocity $\velTn$, and position $\pointTn$ for each material point
    $p=1,\ldots,\nparticles$ at time instance $t_n$}
\KwResult{Deformation gradient $\FTnp$, velocity
    $\velTnp$, and position $\x^p_{n+1}$, $p=1,\ldots,\nparticles$ at time instance $t_{n+1}$}
 
 Transfer Lagrangian kinematics to the Eulerian grid by performing a `particle
    to grid' transfer: Compute for $i=1,\ldots,\nBasis$
 \begin{align*}
    m_{i,n} &= \sumParticles N_i( \pointTn ) \mass \\
    m_{i,n}\vb_{i,n} &= \sumParticles N_i( \pointTn ) \mass \velTn \\
\fb^{\cs}_{i,n} &= -\sumParticles \frac{J(  \FTn ) }{ \rho_0 } \cs( \FTn)\nabla_{\x}  N_i( \pointTn ) \; \mass \\
\fb^{e}_{i,n} &= \sumParticles\frac{J(  \FTn ) }{ \rho_0 } \bb(\pointTn ) N_i( \pointTn ) \; \mass 
\end{align*}

Solve Eulerian governing equations by computing for $i=1,\ldots,\nBasis$
\begin{align*}
    \dot\vb_{i,n+1}  &= \frac{1}{ m_{i,n} } ( \fb^{\cs}_{i,n} + \fb^{e}_{i,n}  ) \\
\Delta \vb_{i,n+1} &= \dot\vb_{i,n+1} \timestepn \\
\vb_{i,n+1} &= \vb_{i,n} + \Delta \vb_{i,n+1} 
\end{align*}

Update the Lagrangian velocity and deformation gradient by performing a `grid
    to particle' transfer: Compute for
    $p=1,\ldots,\nparticles$
\begin{align*}
    \velTnp &= \sumBasis N_i(\x_n^p) \vb_{i, n+1}  \\
\FTnp &= (\1 + \sumBasis \vb_{i, n+1} \otimes \nabla_\x N_i( \x_n^p )
    \timestepn ) \FTn
\end{align*}

Update Lagrangian positions for $p=1,\ldots,\nparticles$
\begin{align*}
    \xb^p_{n+1} &= \xb^p_{n} + \Delta t \velTnp
\end{align*}

 \caption{MPM Algorithm}
    \label{alg:MPM}
\end{algorithm}

\section{Reduced-order model}
\label{sec:rom}
We now propose a methodology for model reduction applicable to the material
point method that relies on constructing a nonlinear approximation to the
deformation map, as well as a family of
projection and hyper-reduction strategies.

\subsection{Kinematics: low-dimensional
manifold}\label{sec:mpkinematics}
In analogue to constructing low-dimensional nonlinear manifolds for
finite-dimensional state spaces \cite{lee2020model}, one can
construct a nonlinear manifold that restricts \textit{any} element of the reference
domain
$\domainRef$ to evolve on a low-dimensional manifold; this can be achieved via an implicit neural
representation. We
first denote the approximated deformation map as $\deformationMapROM:\domainRef\times
\timeInterval\times \paramDomain\rightarrow \RR{d}$
with 
$\deformationMapROM(\cdot;\cdot,\params)\in\Sc(\params)$, $\forall
\params\in\paramDomain$
and \MMC{ Should we use a different term than $\Sc$? Isn't $\deformationMapROM
\in \tilde \Sc \subseteq \Sc$ ? Why the dependance on $\mub$ ?  } \KTC{MC: I'm
not sure why we'd need to do this; by saying we are in $\mathcal S$ we are
just enforcing the initial conditions and boundary conditions exactly. We are
reintroducing $\params$ because this is more important for the ROM
formulation. I think we should keep as is.} \PYC{@MC@KC} \KTC{I'm good with
how it's written now.}
\begin{align}
    \deformationMapROMArgs{\cdot}{t}{\params}\,:\,&\positionRef\mapsto\rev{\positionArg{t}{\params}}\\
    \,:\,&\domainRef\rightarrow \domainROMArgs{t}{\params}\subseteq\RR{d},
\end{align}
where
$\domainROMArgs{t}{\params}\subseteq\RR{d}$ denotes the deformed domain
corresponding to the approximated solution at time
$t\in\timeInterval$ and parameter instance
$\params\in\paramDomain$,
and enforce the kinematic constraint
\begin{equation}\label{eq:manifoldDecoder}
    \deformationMapROMArgs{\X}{\cdot}{\cdot} \in\manifoldX\defeq\{
        \deformationMapApproxArgs{\X}{\genCoordsDummy}\,|\,\genCoordsDummy\in\RR{\nred}\}\subseteq\RR{d},\quad\forall
    \X\in\domainRef,
\end{equation}
where $\deformationMapApprox:\domainRef\times \RR{\nred}\rightarrow \RR{d}$
denotes a parameterization function for a low-dimensional manifold of
dimension $\nred(\ll
\nParticles)$. \MMC{I would specify $\nred \ll
\nParticles$} \PYC{done.} \KTC{Folded it into the writing a bit better}

\begin{figure}[H]
    \centering
    \includegraphics[width=\linewidth]{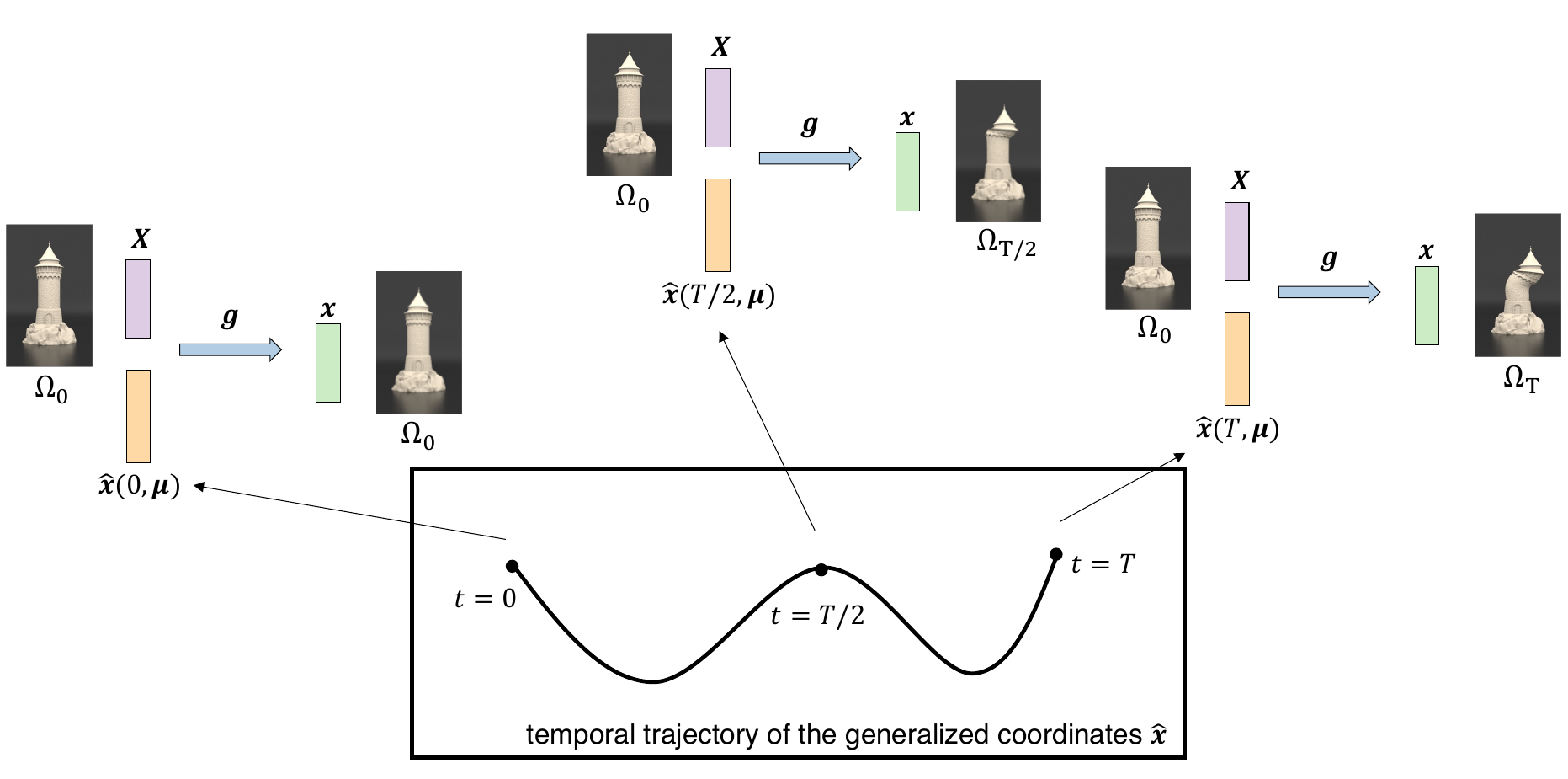}
    \caption{Manifold-parameterization function $\deformationMapApprox$ maps
		the undeformed position $\positionRef$ and the generalized
		coordinates $\genCoords$ to the deformed position \rev{$\position$}.
		We can interpret this approximation as an implicit neural
		representation, as an input argument is the (continuous) domain of the
		function, and the mapping will be learned using neural networks as
		described in Section \ref{sec:manifold-construct}.
    }
    \label{img:manifold-parameterization}
\end{figure}

Pragmatically, the kinematic restriction~\eqref{eq:manifoldDecoder} implies
that there exist generalized coordinates $\genCoords:\timeInterval\times
\paramDomain\rightarrow\RR{\nred}$ such that
 \begin{equation} \label{eq:kinematicRestriction}
\deformationMapROMArgs{\positionRef}{t}{\params} = 
     \deformationMapApproxArgs{\positionRef}{\genCoordsArgs{t}{\params}},\quad\forall\X\in\domainRef,\ \forall 
     t\in\timeInterval,\ \params\in\paramDomain.
 \end{equation}
\Cref{img:manifold-parameterization} schematically illustrates the
manifold-parameterization function $\deformationMapApprox$ underpinning the
proposed kinematic constraint.

 Assuming continuous differentiability of the parameterization function, Eq.~\eqref{eq:kinematicRestriction} implies that the deformation
 gradient of the approximate solution can be calculated analytically as
\begin{align}\label{eq:defGrad}
    \begin{split}
    \deformationGradientApprox\,:&\,(\positionRefDummy,\timeDummy,\params )\mapsto\nabla
 \deformationMapROMArgs{\positionRefDummy}{\timeDummy}{\params}\equiv\nabla \deformationMapApproxArgs{\positionRef}{\genCoordsArgs{t}{\params}}\\
    :&\,\domainRef\times \timeInterval\times
    \paramDomain\rightarrow \RR{d\times d},
    \end{split}
\end{align}
where $\nabla( \cdot) \equiv \frac{\partial}{\partial \positionRefDummy}
(\cdot)$ denotes the gradient with respect to the undeformed position, and
that the velocity of the
approximated solution can be calculated as
\begin{align}\label{eq:timederOne}
        \dot{\deformationMapROM}({\X};t,\params)&\equiv\frac{\partial
        \deformationMapApprox}{\partial
        \genCoords}(\X;\genCoordsArgs{t}{\params})\dot \genCoords(t,\params),\quad\forall\X\in\domainRef,\ \forall 
     t\in\timeInterval,\ \params\in\paramDomain.
\end{align}
where $\dot \genCoords(t,\params)$ denotes the generalized velocity.

Recall that we require the approximated deformation map to reside in the space
of admissible trajectories such that 
$\deformationMapROM(\cdot;\cdot,\params)\in\Sc(\params)$, $\forall
\params\in\paramDomain$. The boundary condition 
$\dot{\deformationMapROM}({\X};t,\params) =
    \overline \Vb(\positionRef, t;\params), \, \forall \positionRef \in \partial_D
    \domainRef(t;\params), \ \forall t\in\timeInterval,\, \forall\params\in\paramDomain$ can be satisfied
    trivially by enforcing the associated boundary conditions to match the
    prescribed ones during the reduced-order simulation.

    \MMC{In the below equation should it be the ROM deformation map or the FOM;
    you are not guaranteed that the ROM will be able to satisfy the equality. I
    would rephrase this paragraph.} \KTC{Peter---We need to re-introduce the expressions I
    put in previously that guaranteed satisfaction of the boundary conditions; I
    sent an email about this. Then you can always include the least-squares
    approach below as an alternative way to approximate the boundary conditions
    that is not exact} \PYC{done.}

To satisfy the initial conditions \begin{align}\label{eq:essential}
    \begin{split}
        \deformationMapROMArgs{\positionRef}{0}{\params} &= \X,\ \forall
				\Xb\in\domainRef(\params),\ \forall \params\in\paramDomain
\\
        \dot{\deformationMapROM}({\X};0,\params) &= \overline
				\Vb(\positionRef,0;\params),\ \forall \Xb\in\domainRef(\params),\
				\forall \params\in\paramDomain,
    \end{split}
\end{align} we represent the manifold-parameterization function as
\begin{equation} \label{eq:defMapApproxForm}
  \deformationMapApprox:(\X,\genCoords)\mapsto
  \deformationMapApproxOne(\X,\genCoords) +
  \deformationMapApproxTwo(\X;\params) +
  \deformationMapApproxThree(\X;\params)\deformationMapApproxFunc(t)
  \end{equation} 
  where 
$\deformationMapApproxOne:\domainRef\times \RR{\nred}\rightarrow \RR{d}$ is the approximated manifold-parameterization function,
$\deformationMapApproxTwo:\domainRef\times\paramDomain\rightarrow\RR{d}$
$\deformationMapApproxThree:\domainRef\times\paramDomain\rightarrow\RR{d}$, and
  $\deformationMapApproxFunc:\timeInterval\rightarrow\RR{}$ satisfies
  $\deformationMapApproxFunc(0) = 0$ and $\dot\deformationMapApproxFunc(0) =
  1$ (e.g., $\deformationMapApproxFunc:t\mapsto t$). Given the functional form
  \eqref{eq:defMapApproxForm}, one can satisfy the initial
  conditions \eqref{eq:essential} at any parameter instance
  $\params\in\paramDomain$ for any prescribed initial values of 
$\genCoords(0,\params)$ and $\dot\genCoords(0,\params)$
  by setting
\begin{align}\label{eq:essentialSpec}
  \begin{split}
      \deformationMapApproxTwo(\X;\params) 
      &= \positionRef-
      \deformationMapApproxOne(\X,\genCoords(0;\params)) 
      ,\ \forall \Xb\in\domainRef,\ \forall \params\in\paramDomain
\\
      \deformationMapApproxThree(\X;\params) 
 &= \overline
      \Vb(\positionRef,0;\params)-
\frac{\partial
      \deformationMapApprox}{\partial
      \genCoords}(\X;\genCoordsArgs{0}{\params})\dot \genCoords(0,\params) ,\
      \forall \Xb\in\domainRef,\ \forall \params\in\paramDomain.
  \end{split}
\end{align}

Additionally, we can obtain a good approximation of the initial boundary
conditions utilizing $\deformationMapApproxOne$ alone by choosing
$\genCoords(0;\params)$ and $\dot \genCoords(0;\params)$ that minimize the
$L^2$-norm of $\deformationMapApproxTwo$ and $\deformationMapApproxThree$
for any $\params\in\paramDomain$, \rev{i.e.,
}\begin{align}\label{eq:essentialSpecInt}
    \begin{split}
			\genCoords(0;\params)\in\underset{\genCoordsDummy}{\text{argmin}}\int_{\domainRef(\params)}\|\X-\deformationMapApproxOne(\X,\genCoordsDummy) \|^2d\Xb
\\
			\dot
			\genCoords(0,\params)\in\underset{\dot\genCoordsDummy}{\text{argmin}}
			\int_{\domainRef(\params)}
\|\overline\Vb(\positionRef,0;\params)-\frac{\partial
        \deformationMapApprox}{\partial
        \genCoords}(\X;\genCoordsArgs{0}{\params})\dot \genCoordsDummy
\|^2d\Xb.
    \end{split}
\end{align}
In practice, we approximate these integrals via numerical quadrature.

\begin{figure}[H]
    \centering
    \includegraphics[width=0.6\linewidth]{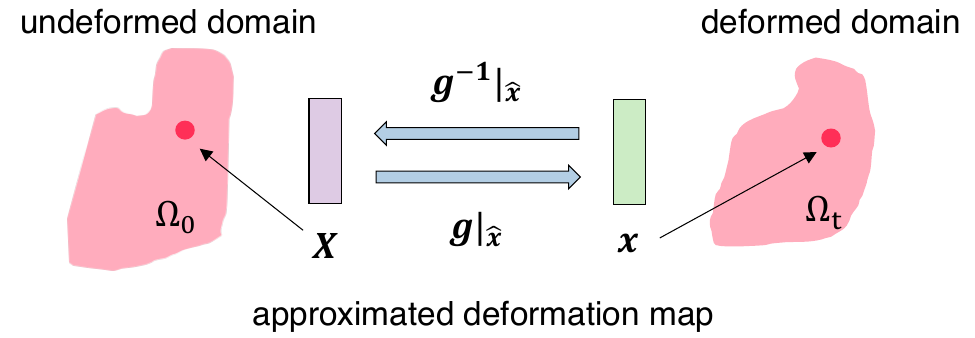}
    \caption{Given the generalized coordinates $\genCoords$, the approximated
		deformation map allows us to recover the current position of an arbitrary
		point from the undeformed domain (Eq. \eqref{eq:kinematicRestriction}). Other kinematic information, such as the deformation gradient and the velocity, can also be recovered via differentiating the approximated
		deformation map (Eqs. \eqref{eq:defGrad} and \eqref{eq:timederOne}). In
		addition, given an arbitrary point from the deformed domain, we can invert
		the approximated deformation map to obtain its undeformed position.\KTC{PC: can you provide a few more requirements on $g$ that ensure
		invertibility per our previous discussions?}\PYC{done.} The approximated deformation map is invertible as long as the approximated deformation map is non-degenerate, i.e., the determinant of the deformation gradient $J$ is nonzero.}
    \label{img:recover-kinematics}
  \end{figure}

\begin{remark}[Recover kinematics of any material point]
We emphasize that---because this approach approximates the entire deformation
	map---given the value of the generalized coordinates $\genCoords(t,\params)$
	and its time derivative $\dot \genCoords(t,\params)$,
	we can compute the
    displacement, the deformation gradient, and the velocity for 
\textit{any element of the reference domain} $\X\in\domainRef$ via
	Eqs.~\eqref{eq:kinematicRestriction}, \eqref{eq:defGrad}, and
	\eqref{eq:timederOne}, respectively (\Cref{img:recover-kinematics}).
	Further, assuming the parameterization function $\deformationMapApprox$ is
	bijective between $\domainRef$ and $\domainROMArgs{t}{\params}$ for a given
	value of the generalized coordinates $\genCoords(t,\params)$, we can even
	invert the approximated deformation map to obtain the undeformed position of
	an arbitrary point in the deformed domain $\domainROMArgs{t}{\params}$.
	\KTC{PC tried to get more precise in previous sentence, thoughts?}\PYC{good.} Consequently, our approach
	supports adaptive quadrature for computing full-space dynamics as well as
	super-resolution. 
    \MMC{I would expand this and have a schematic explaining it} \KTC{Agree with
    MC. We also need to include an explanation for inverting the deformation map
    to discover the point in $\Omega_{ref}$ corresponding to the desired
    quadrature points in $\Omega$.} \KTC{I agree with MC; we should be more
    precise in the descriptions; this reads too high-level currently} \PYC{done.}
    
\end{remark}

\subsection{Dynamics}
\label{sec:dynamics}
\begin{figure}[H]
    \centering
    \includegraphics[width=0.7\linewidth]{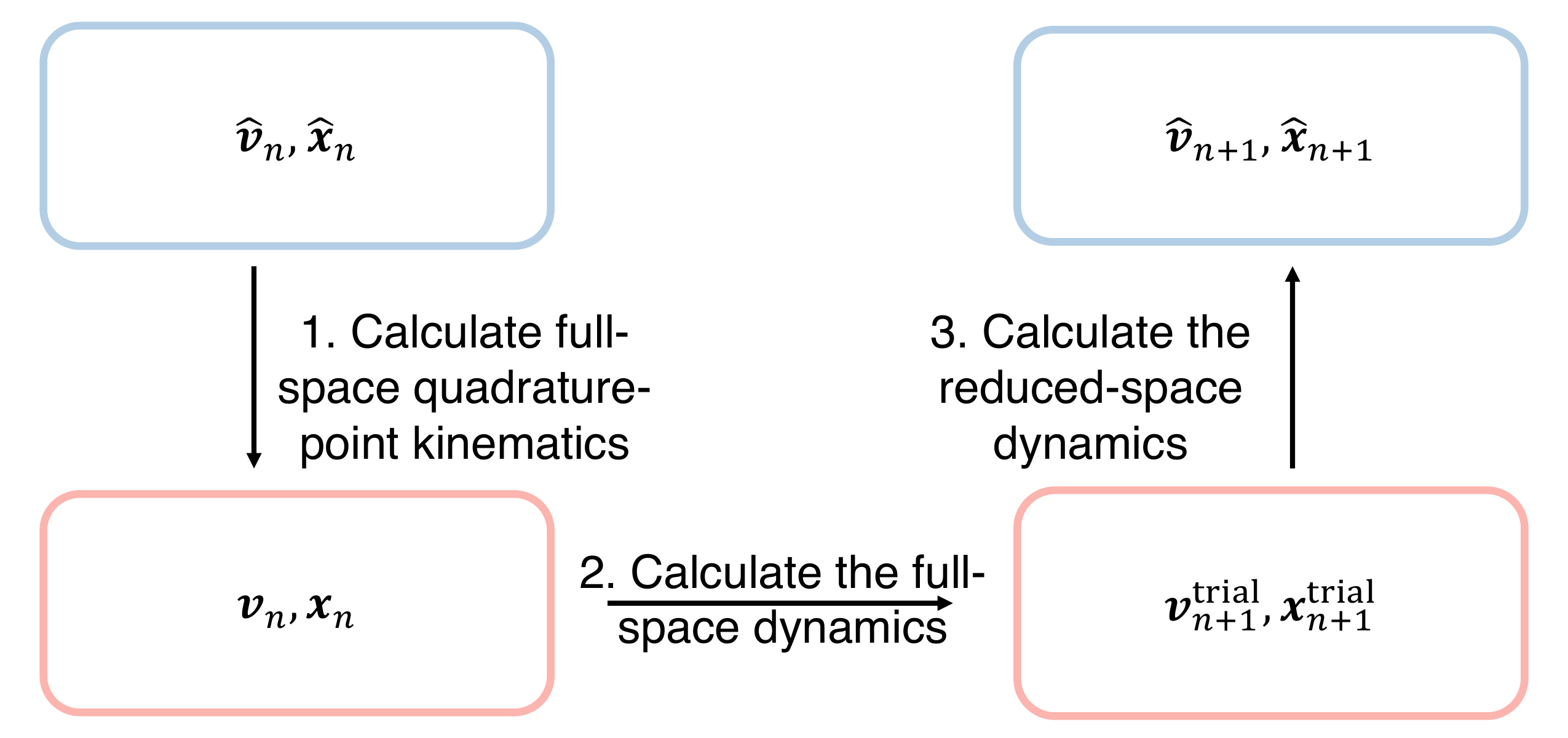}
    \caption{Reduced-order dynamics (Algorithm
        \ref{algo:reduced_order_dynamics}) \MMC{(0)Fonts are a complete
				mismatch}\KTC{(1)reflect the description in the abstract (2)
				`trial'unitalicized}}\PYC{(0) helvetica throughout; (1)and(2) done.}
				\KTC{PYC: can you typeset the v trial and x trial a bit tighter? see
				the algorithms for how I fixed this elsewhere} \KTC{PYC: can you also
				rename the first step `Calculate full-space quadrature-point
				kinematics' or similar to emphasize quadrature points?}\PYC{done.}
    \label{img:algo-flow-chart}
  \end{figure}
We compute the dynamics needed to evolve the generalized coordinates and
velocity in three steps (see \Cref{img:algo-flow-chart}):\KTC{rephrase this
and the caption of the figure to mirror the description in the abstract; these
three steps should be described consistently throughout the paper.}\PYC{done.}
Calculate full-space kinematics at quadrature points (\Cref{sec:quadratures}),
Calculate the full-space dynamics for a small number of `sample material
points' (\Cref{sec:evolution}), and Calculate the
reduced-space dynamics (\Cref{sec:projection}). The selection of sample material points is described in Section \ref{sec:hyper-reduction}.

Algorithm \ref{algo:reduced_order_dynamics} presents the complete
algorithm for generating reduced-order dynamics. We note that displacement and
velocity in the full space can always be decoded from the generalized
coordinates, e.g., for rendering, computing quantities of interest.

\begin{algorithm}[H]
    \SetAlgoLined
    \KwData{Generalized velocity $\genVels_{n}$, generalized coordinates $\genCoords_n$.}
    \KwResult{Generalized velocity $\genVels_{n+1}$, generalized coordinates $\genCoords_{n+1}$.}
    
    Calculate full-space kinematics at quadrature points via either Algorithm \ref{algo:evolution-qmpm} or Algorithm \ref{algo:evolution-qfem}.

    Calculate the full-space dynamics for sample material points to obtain ${\velTnpTrial}$ and $
	\xTnpTrial$ for $p\in\sampleSet$\KTC{note I had to change the macro for
	sampleSet to $\mathcal P$ because we double-used $\mathcal M$ for the
	manifold and these sample points.} using Algorithm \ref{algo:mpm-style-dynamics}.
    
    Calculate the reduced-space dynamics by projecting ${\velTnpTrial}$ and $
\xTnpTrial$
		onto the reduced space via either Algorithm \ref{algo:projection_nonlinear} or Algorithm \ref{algo:projection_linear}.
    
        \caption{Reduced-order-model dynamics \KTC{Describe these three steps
        consistently throughout the paper}}
    \label{algo:reduced_order_dynamics}
\end{algorithm}

\subsubsection{Hyper-reduction}
\label{sec:hyper-reduction}
\KTC{I would actually put this as a subsection before the current subsection 3.2.1, as knowing about this set of
points $\mathcal M$ is necessary for all of 3.2.1--3.2.3.} 
\MMC{I would be more precise on
this}
\KTC{Agree with MC; start with the last step (need to project
position/velocity at only a few points to keep the least-squares problem
overdetermined) and work backward (because of this, only need to compute
position/velocity at a few points; because of this, only need kinematics
around those points}
\KTC{I recommend moving Section \ref{sec:hyper-reduction} here, and changing
it to be a subsubsection. The next three subsubsections all require knowing
about the set of points $\sampleSet$.}
\PYC{moved this section to the front as suggested. my only concern is now we will mention the next three sections without going to the details in order to explain why we need a set M.}

A common theme across all three steps in the reduced-order dynamics
(\Cref{img:algo-flow-chart} and Algorithm \ref{algo:reduced_order_dynamics})
is that only a subset of the original material points is required for
computation. This opportunity arises from Step 3: because we only need to
update a small number (i.e., $\nred\ll\nParticles$) of generalized
coordinates, we can drastically undersample the full-space kinematics, yet
retain an overdetermined least-squares problem for this dynamics projection.
As such, we achieve significant computational-cost savings by performing this
projection using
a small subset of the original material points, which we refer to as the
`sample material points' indexed by 
$\sampleSet\subseteq\{1,\ldots,\nParticles\}$, where
$\frac{\nred}{d}\leq|\sampleSet|\ll\nParticles$. This ensures the
reduced-order simulation incurs $\nParticles$-independent
computational complexity; the set of approaches that enable reduced-order
models to operate on a small subset of thee domain
is often referred to as hyper-reduction in the literature
\citep{ryckelynck2005priori}.
\KTC{I'd rephrase this
argument. I'd say something like `because the least-squares problems appearing
in Algorithm \ref{algo:projection_nonlinear} require computing only
$\nred(\ll\nParticles)$ unknowns, we can sample far fewer than $\nParticles$
residual equations, yet retain overdetermined least-squares problems as long 
as $|\sampleSet|\geq \nred$.}\PYC{done.} \KTC{PC: don't we need
$\frac{1}{d}\nred\leq|\sampleSet|$ because the least squares problem is
algebraic and there are $d\times |\sampleSet|$ equations (one for each DOF of
each sample material point) constraints in the least squares problem?}\PYC{yes!}

As a consequence of employing a small number of material points in Step 3, the
second step of Algorithm
\ref{algo:reduced_order_dynamics} also only requires computing the dynamics
for the small number of sample material points belonging to the set $\sampleSet$. To
calculate these dynamics updates, Step 1 of Algorithm \ref{algo:reduced_order_dynamics} requires computing 
kinematic information only at quadrature points that share Eulerian
basis-function support with the sample material points.

While more advanced methods for choosing sample material points $\sampleSet$ exist
\citep{an2008optimizing,carlberg2011efficient}, we adopt a straightforward
stochastic sampling scheme due to its simplicity, wherein we 
re-sample at every time step to ensure good
coverage of the domain.\KTC{This previous sentence implies challenges
for the Lagrangian-quadrature kinematics; how do we track all of these
neighbors? We should reconsider this or explain it carefully.}\PYC{tracking neighboring points is discussed in detail in the lagrangian vs. eulerian section. I don't see why we need to discuss it again here. the sampling approach applies to both cases.} 
\KTC{I guess I'm still confused: is re-sampling even possible with Lagrangian
quadrature? If not, be more careful in how you describe this for Lagrangian v
Eulerian quadrature}\PYC{later lagrangian vs. eulerian sections clarified this. here we just discussing sampling regarding projection, not kinematic point}If kinematic boundaries exist, special
attention is given to them by ensuring that the material points from
these boundaries are included in $\sampleSet$.\KTC{Previous sentence: what
requirements do we impose to ensure solvability? This is not currently
rigorous. I believe that at the number of required
Dirichlet-boundary-condition samples is equal to the dimension of the
generative process for all possible boundary conditions. Further, those
samples must ensure invertibility of the full underlying boundary condition.
Either way, we need stronger, precise statements here.}\PYC{added a sentence after this comment. This is not a hard constraint during projection tho. Even if we pose it as a hard constraint, like you said, it is not necessarily solvable and can be enforced only in the least-sqaures sense. there is quite a bit of recent work on exact boundary condition.} In particular, these kinematic material points' Dirichlet boundary conditions are strictly enforced during the full-space update (\Cref{sec:evolution}). 

\begin{figure}
    \centering
    \includegraphics[width=1.0\textwidth]{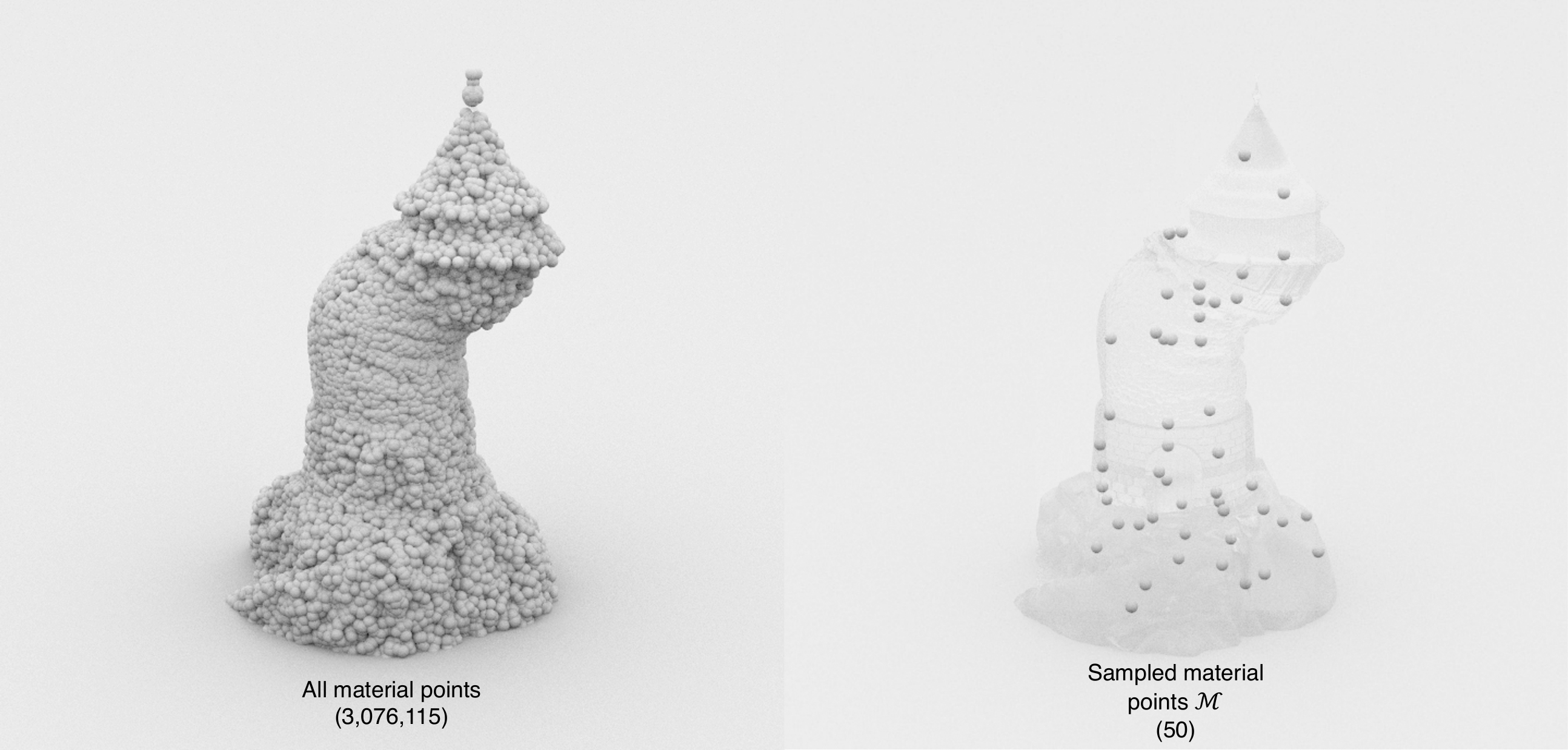}
    \caption{The sample material points $\sampleSet$ are shown on the right. Note that both the top and the bottom kinematic boundaries are sampled. \MMC{This picture can be improved starting with the mismatching font to larger point sizes and perhaps with the surface mesh superimposed with low opacity.} }\PYC{font size aimed at matching the rest of the font size. Done.}
    \label{img:hyperreduction:visualize} \MMC{Love this new figure! Great work!}
  \end{figure}

\Cref{img:hyperreduction:visualize} displays an example of the sample material
point set
$\sampleSet$ from one of the experiments that will be discussed in Section
\ref{sec:experiments}. \KTC{Added the last part of the sentence to ground it to something
that's coming later.}

\subsubsection{Calculate full-space kinematics at quadrature points}
\label{sec:quadratures}
\KTC{I recommend carefully rewriting this section with the comments below in
mind. There were quite a few things that needed substantial cleaning up.}

\KTC{ Also
note that we immediately introduce $\sampleSet$ without first explaining it,
which is done in Section \ref{sec:hyper-redution}. As such, we should move
that section to appear before this one.}
\PYC{done}

To compute the dynamics of the sample material points
$\sampleSet\subseteq\{1,\ldots,\nParticles\}$, the equation of motion has to be integrated \eqref{eq:massWeakForm}. We can discretize the spatial domain \eqref{eqn:material_point} with either Lagrangian quadrature points defined by the reference configuration or Eulerian quadrature points defined by the current configuration. Consequently, in the process of numerically evaluating the integral \eqref{eq:weak_form_discrete_space}, we can use a quadrature rule defined either on the
Lagrangian quadrature points or the Eulerian
quadrature points. Note that the original MPM algorithm adopts the Lagrangian quadrature approach where the material points serve as the Lagrangian quadrature points.\MMC{I would
talk about the fact that we can compute the integral into the two different
configurations, the reference and the current configuration, which can be
approximated with a quadrature rule on the reference domain (ie. the
Lagrangian quadrature) or one on the current domain (ie. the Eulerian
quadrature), respectively}\PYC{done} \KTC{I find this wording a bit confusing
because we are still computing quadrature rules for the \textit{Eulerian}
governing equations. Perhaps think of a way to make the argument a bit more
clear.}\PYC{edited. MMC's previous comment is incorrect on second thoughts. our integral is always evaluated on the current configuration. however, the quadrature can have either lagrangian or eulerian origin.}

\PYC{put the quadrature definition up front. instead of Eulerian only.}

\KTC{Peter, shouldn't we be referring to Eqs (15) and (16) here? Maurizio
introduced this type of discretization already}\PYC{worth restating b/c the notion is slightly different.you added a wondeful sentence at the end of the paragraph.} Formally, we can discretize the weak form \eqref{eq:massWeakForm} using quadrature points:
\begin{equation} 
    \sumQuads (  \sumjBasis \dot \vb_j N_j \, N_i )|_{\quadPoint} \;  \quadMass = 
        \sumQuads \frac{1}{\rho_0}\left[ J \; (  \bb N_i - \cs( \Fb ) \nabla N_i )
        \right]|_{\quadPoint} \quadMass + \int_{\partial_N
        \Omega} \overline \tb N_i, \quad
        i=1,\ldots,\nBasis
        \label{eq:weak_form_discrete_space_quad} .
    \end{equation}
where $\nQuad$ denotes the number of the quadrature points and the left
superscript $(\cdot)^{Q}$ indicates quantities related to the quadrature
points. Note that Eq.~\eqref{eq:weak_form_discrete_space_quad} generalizes
Eq.~\eqref{eq:weak_form_discrete_space} from the original material point
method to allow for hyper-reduction and alternative quadrature rules. \KTC{added the
last sentence here to draw the connection to the original similar form in
MPM.}

\begin{figure}[H]
    \centering
    \includegraphics[width=\linewidth]{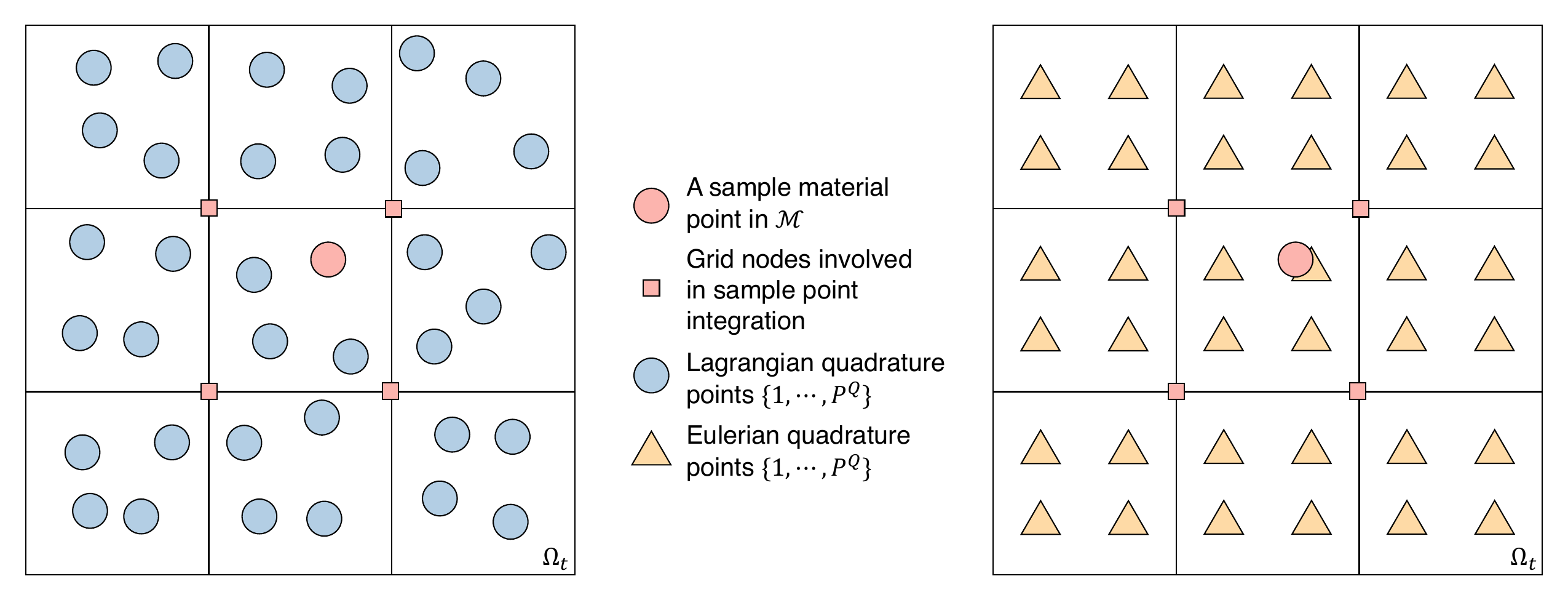}
    \caption{Lagrangian quadrature points (left) vs. Eulerian quadrature
        points (right). All depicted
        quadrature points are required to compute the dynamics for the depicted
        sample material point. Note that in the Lagrangian quadrature approach, the sample material point itself will also serve as a quadrature point, which is not the case with the Eulerian quadrature approach.\KTC{use $\mathcal N$ in the legend for
        both the Lagrangian and Eulerian Quadrature points.}\PYC{done.}\KTC{Also, in the rightmost
        figure, you're missing the top-right Eulerian quadrature point in the
				middle cell.}\PYC{done.}\KTC{I'd clarify that the domain shown here
				is the \textit{Eulerian domain} $\deformationMapROM(\cdot;t,\params)$
				in both cases.}\PYC{done.} Both domains shown are the deformed domain $\domainDef$. }
    \label{img:qmpmvsqfem}
  \end{figure}

\myparagraph{Quadrature points via Lagrangian material points}

The first approach is similar to that employed by the original MPM
algorithm. Here, we use the neighboring material points to define the
quadrature rule
(\Cref{img:qmpmvsqfem} left).
Formally, we consider the neighbors of the sample material points $\neighborSet\subseteq\{1,\ldots,\nParticles\}$,
which we define as the subset of all material points that share Eulerian basis-function
support with the sample material points (with
$\sampleSet\cap\neighborSet=\emptyset$).\KTC{should probably clarify that these
are the material points characterizing the full-order-model discretization}\PYC{done.} \KTC{we'll also need to clarify the actual quadrature
points as $x^i$, $i\in\mathcal P = \mathcal M\cup\mathcal N$}\PYC{done.}
Together, the sample material points and the neighboring material points form
the set of the quadrature points, i.e., $\quadPointTn=\pointTnPi$, $p\inquadPointSet$, where $\nQuad = |\sampleSet\cup\neighborSet|$ and $\quadmap: \quadPointSet\rightarrow \sampleSet\cup\neighborSet$ is a bijective mapping between the two sets.

Algorithm \ref{algo:evolution-qmpm} provides the
algorithm that identifies these quadrature points and obtains their kinematic information from the approximated deformation map.

\begin{algorithm}[H]
    \SetAlgoLined
    \KwData{Generalized velocity $\genVels_{n}$, generalized coordinates $\genCoords_{n}$.}
    \KwResult{Quadrature-point
        kinematics $\quadMassTn$, $\quadPointTn$, $\quadVelTn$, $\quadFTn$ for $p\inquadPointSet$}

    Compute the position
        $\pointTn$ for each sample material point
        $p\in\sampleSet$ at time instance $t_n$ by evaluating
        \eqref{eq:kinematicRestriction} for
        $\X=\X^p$, ${p\in\sampleSet}$ and $t=t_n$.

    Identify the basis functions $\basisFunctionSet$ needed to compute dynamics for the sample
    material points, i.e., $\basisFunctionSet= \{i
\in\{1,\ldots,\nBasis\}
		\,|\,\exists p\in\sampleSet\ \text{s.t.}\ N_i(
    \pointTn)\neq 0\}
		$.

    Identify the neighbor material points set $\neighborSet$, i.e., 
	$\neighborSet = \{p\in\{1,\ldots,\nParticles\}\setminus\sampleSet
	\ |\ \exists
	i\in\basisFunctionSet\ \text{s.t.}\
	N_i(\deformationMapApprox(\X^p;\genCoords_{n}))\neq 0\}$.

    Compute the deformation gradient $\FTn$ and the velocity $\velTn$ for each sample and neighbor material point at time instance $t_n$ by evaluating \eqref{eq:defGrad} and \eqref{eq:timederOne} for $\X=\X^p$, ${p\in\sampleSet\cup\neighborSet}$, and $t=t_n$. 
 
    Set the quadrature point kinematics to be $\quadMass=\massPi$, $\quadPointTn=\pointTnPi$, $\quadVelTn=\velTnPi$, $\quadFTn=\FTnPi$ for $p\inquadPointSet$
    , where $\nQuad = |\sampleSet\cup\neighborSet|$ and $\quadmap: \quadPointSet\rightarrow \sampleSet\cup\neighborSet$ is a bijective mapping between the two sets.

    \caption{Full-space kinematics at Lagrangian quadrature points (via tracking material points)}
    \label{algo:evolution-qmpm}
\end{algorithm}

This approach is most similar to the original emulated MPM approach, and it can incur an operation count independent of the original number of material points $\nparticles$ and Eulerian basis functions
$\nBasis$. However, it does require computing (and tracking) the set of neighboring
material points. In the worst case, tracking would result in
$\nparticles$-\rev{dependent} complexity due to the
difficulty of ascertaining \textit{a priori} the set of neighboring material points that
will ever be encountered for a targeted sample point for any possible online
trajectory. \MMC{I would add this remark in the previous
section}\PYC{done} \KTC{Agree. In the previous section, should also add the benefit,
i.e., ``While this approach is most similar to the original emulated MPM
approach, it can incur worst-case complexity of $O(\nparticles)$ due to the
difficulty of ascertaining \textit{a priori} the set of neighboring material points that
will ever be encountered for a targeted sample point for any possible online
trajetory''}\PYC{done}

\myparagraph{Quadrature points via Eulerian quadrature points\KTC{YC: why
`finite elements'? This is more just typical numerical quadrature, finite
elements implies a lot more machinery}\PYC{agree!}}

\KTC{Notation is broken for this approach. We have already denoted $x^p$,
$p=1,\ldots,\nParticles$ to be
the \textit{original} set of material points. In Eulerian quadrature, we are
inventing new points not present in the original set. Just calling them $x^q$
is not sufficient, because the superscript is just a dummy variable; we could
just as equivalently refer to the original materal points as 
$x^i$,
$i=1,\ldots,\nParticles$ or 
$x^j$,
$j=1,\ldots,\nParticles$ or
$x^q$,
$q=1,\ldots,\nParticles$.
Maybe we refer to these points as $y^p$ (or $y^i$, or $y_j$, ...).
}
\PYC{Not only positions, but all the kinematic information needs to be differentiated from the original MPM discretization. In addition, Lagrangian and Eulerian quarature should use the same notations because they are all \emph{quaratures}. They are utilized the same way in the next section during the full-space update. Therfore, we have switched using subscript $^Q$ for quadrature quantities.}

\KTC{, whereas here,
the quadrature point set would be $y^i$, $i=1,\ldots \nQuad$.} \PYC{after notation change, lag and eul has the same quarature point set notation, which are differentiated from the original MPM discretization.}

\KTC{I'm not
sure why we are discussing it this way, because the previous section just
fleshed this out in its entirety! I'd conclude the previous section with a
discussion of this drawback to motivate this section}\PYC{done. previous discussion removed.}

To avoid the costly tracking of these neighboring material points, we present an alternative that generates Eulerian quadrature points instead of the Lagrangian quadrature points (\Cref{img:qmpmvsqfem} right). The resulting
quadrature rules discretize the Eulerian configuration instead of the Lagrangian configuration.\KTC{previous sentence doesn't
make sense to me}\PYC{edited.} We generate $\ell $ quadrature points per dimension and per background grid cell.\KTC{In previous
sentence, we never start a sentence with a mathematical symbol}\PYC{edited.} These quadrature point locations are evenly distributed such that the distance between each pair of neighboring quadrature point is $\frac{\cw}{\ell }$, where $\cw$ is the grid-cell width. Therefore, for each
quadrature point, we have its current position $\quadPoint$ and its volume
$\quadVol=\frac{1}{\ell ^d}\cellVol$, where $\cellVol=(\cw)^d$ denotes the volume of the grid cell. \KTC{How
do we compute the volume of the grid cell? This seems highly
non-obvious.}\PYC{edited.}\KTC{This last sentence is imprecise, and won't fly
in JCP. What rigorously can be said about this? Otherwise, I'd just leave
$\ell=2$ as a choice mentioned in the experiments, not here.}\PYC{removed.}

To compute the undeformed position of the Eulerian quadrature points in the
reference configuration, we can invert
the approximated deformation by computing $\quadPointIni$ such that
$\deformationMapApprox(\quadPointIni;\genCoords)=\quadPoint$. Other kinematic
quantities can then be computed using \Cref{eq:defGrad} and
\Cref{eq:timederOne}. The mass $\quadMass$ of each quadrature point can be computed
as $\quadMass=\frac{\rho_0}{\quadJTn}\quadVol$.\KTC{this doesn't explain how to compute the
$x^q$ in the first place, though}\PYC{added defn in the prev paragraph} These Eulerian quadrature points can then be used the same way as the Lagrangian quadrature points to discretize the weak form of the equation of motion \eqref{eq:weak_form_discrete_space_quad}.

Algorithm \ref{algo:evolution-qfem} presents the associated algorithm. In
contrast to the first approach, this approach does not require tracking any
neighboring material points. Instead, leveraging the invertibility of the
deformation map, we can generate quadrature points necessary for updating the
dynamics of the sample material points.\KTC{This is the first
time $\basisFunctionSet$ has appeared within the prose. I'd recommend careful
rewriting of Section 3.2 with all these comments in mind.}\PYC{remove uncessary mention of $\basisFunctionSet$} Consequently, the
approach can incur an operation count independent of the original number of
material points $\nparticles$ and Eulerian basis functions
$\nBasis$ without any additional tracking.

\begin{algorithm}[H]
    \SetAlgoLined
    \KwData{Generalized velocity $\genVels_{n}$, generalized coordinates $\genCoords_{n}$.}
    \KwResult{Quadrature-point
        kinematics $\quadMass$, $\quadPointTn$, $\quadVelTn$, $\quadFTn$, $p\inquadPointSet$}

    Compute the position
        $\pointTn$ for each sample material point
        $p\in\sampleSet$ at time instance $t_n$ by evaluating
        \eqref{eq:kinematicRestriction} for
        $\X=\X^p$, ${p\in\sampleSet}$ and $t=t_n$.

    Identify the basis functions $\basisFunctionSet$ needed to compute dynamics for the sample
    material points, i.e., $\basisFunctionSet=
		\{i\in\{1,\ldots,\nBasis\}\,|\,\exists p\in\sampleSet\ \text{s.t.}\ N_i(
    \pointTn)\neq 0\}$.

        \KTC{Notation is broken for this approach. Changing the label of a dummy
        variable (supersript) doesn't change the identity of the variable. We
        should use a different fundamental notation (e.g., y instead of x) for
        invented material points, as $x$ with a superscript already denotes the
        set of original MPM material points.}\PYC{new notation defined in the main text} Define a quadrature rule comprising quadrature points and their volumes
        \KTC{need to clarify how the volumes are computed, this is not obvious}\PYC{claried in the main text}
    $\quadPointTn\in\Omega$, $\quadVol\in\RR{}_+$, $p\inquadPointSet$, used to
    assemble the governing equations at the sample nodes $\basisFunctionSet$.
		Compute the undeformed positions of the quadrature points
		$\quadPointIniTn$ by solving
		$\deformationMapApprox(\quadPointIniTn;\genCoords_{n})=\quadPointTn$,
		$p\inquadPointSet$.

    Compute the deformation gradient $\quadFTn$ and the velocity $\quadVelTn$ for each quadrature point at time instance $t_n$ by evaluating \eqref{eq:defGrad} and \eqref{eq:timederOne} for $\X=\quadPointIniTn$ and $t=t_n$.

    Compute the mass $\quadMass$ for each quadrature point, $\quadMass=\frac{\rho_0}{\quadJTn}\quadVol$ where $\quadJTn = \det(\quadFTn)$.

		\caption{Full-space kinematics at Eulerian quadrature points (via
		inverting the deformation-map approximation)}
    \label{algo:evolution-qfem}
\end{algorithm}

\begin{remark}
        In addition to reducing the computational cost, the ability to generate
        arbitrary quadratures also enables adaptive refinement, which can be
        instrumental when there is extreme deformation (i.e., the determinant of the deformation gradient is large\KTC{What does it mean for the `Jaobian tgo get large'? Do you mean
        `the Jacobian determinant becomes large'?}\PYC{fixed}).
    \MMC{We need to make a remark about the transfer of internal state
        variables; the reviewer will 100\% comment on this.}  \KTC{MC: something
        like `this work considers only elasticity; for plasticity, we will need to
        equip the approach with a mechanism to predict fields of internal state
        variables'?}\PYC{done.}
\end{remark}
\begin{remark}
    Since this work considers only elasticity, the quadrature points do not carry internal state variables; for plasticity, we will need to
				equip the approach with a mechanism to predict (continuous) fields of internal state variables. These fields could also be pursued with the mesh-independent low-dimensional manifold presented in Section \ref{sec:mpkinematics}.
\end{remark}

\subsubsection{Calculate the full-space dynamics \KTC{Don't like this title, should
be consistent with the framing in the abstract and elsewhere}\PYC{done}}
\label{sec:evolution}
This section calculates the full-space dynamics by computing the trial velocities and positions for the material points
belonging to the subset $p\in\sampleSet$ at $t^{n+1}$.\KTC{Two things: (1)
please refer to this as computing full-space dynamics, as this is done
elsewhere; (2) we also compute the trial velocities, not sure why this isn't
mentioned}\PYC{done.} We deem these velocities and positions
\emph{\text{trial}} because they do not necessarily respect the kinematic
constraint \eqref{eq:manifoldDecoder} induced by the low-dimensional manifold; they are simply the updates to the velocities and
positions that would be computed from the current state by the full-order
model, restricted to the set of sample material points.\KTC{I added the last
part of the previous sentence to clarify, please verify}\PYC{done.} Algorithm \ref{algo:mpm-style-dynamics} presents the
MPM-style dynamics calculation that works for both the Lagrangian-quadrature kinematics and the
Eulerian-quadrature kinematics\KTC{again, would prefer the more precise terms
`Lagrangian-quadrature kinematics and Eulerian-quadrature kinematics'.}\PYC{done}.

\begin{remark}
A salient difference between the original MPM algorithm
(Algorithm \ref{alg:MPM}) and the dynamics calculation presented here is that this new approach no longer needs
to evolve the deformation gradient explicitly. The deformation gradient is
readily available from the approximated deformation
	map's spatial gradient as derived in Eq.~\eqref{eq:defGrad}. \KTC{I'd consider separating this fact---the lack of need to explicitly
update deformation gradient---into a remark.}
\end{remark}\PYC{done.}

\begin{algorithm}[H]
    \SetAlgoLined
    \KwData{Quadrature-point kinematics $\quadMass$, $\quadPointTn$, $\quadVelTn$, $\quadFTn$, $p\inquadPointSet$}
    \KwResult{Full-space trial positions $\xTnpTrial$ and velocities ${\velTnpTrial}$ for $p\in\sampleSet$.}
 
 Perform the `particle to grid' transfer by computing for
    $i\in\basisFunctionSet$
 \begin{align*}
    m_{i,n} &= \sumQuads N_i( \quadPointTn ) \quadMass \\
    m_{i,n}\vb_{i,n} &= \sumQuads N_i( \quadPointTn ) \quadMass \quadVelTn \\
\fb^{\cs}_{i,n} &= -\sumQuads \frac{J(  \quadFTn ) }{ \rho_0 } \cs( \quadFTn)\nabla_{\x}  N_i( \quadPointTn ) \; \quadMass \\
\fb^{e}_{i,n} &= \sumQuads\frac{J(  \quadFTn ) }{ \rho_0 } \bb(\quadPointTn
     ) N_i( \quadPointTn ) \; \quadMass. 
\end{align*}

Perform the update step by computing for $i\in\basisFunctionSet$
\begin{align*}
    \dot\vb_{i}  &= \frac{1}{ m_i } ( \fb^{\cs}_{i,n} + \fb^{e}_{i,n}  ) \\
\Delta \vb_{i} &= \dot\vb_{i} \timestepn \\
\vb_{i,n+1} &= \vb_{i,n} + \Delta \vb_{i}.
\end{align*}

Perform the `grid to particle' transfer by computing for
    $p\in\sampleSet$
\begin{align*}
    {\velTnpTrial} &= \sumBasisFunctionSet N_i(\x_n^p) \vb_{i, n+1}
\end{align*}

Update Lagrangian positions for $p\in\sampleSet$
\begin{align*}
	\xTnpTrial &= \xb^p_{n} + \Delta t {\velTnpTrial}
\end{align*}

    \caption{Full-space dynamics for sample material points}
    \label{algo:mpm-style-dynamics}
\end{algorithm}

\subsubsection{Calculate the
reduced-space dynamics \KTC{Don't like this title, should
be consistent with the framing in the abstract and elsewhere, e.g., `calculate
reduced-space dynamics'}\PYC{done}}
\label{sec:projection}
This section proposes two approaches for computing reduced-space dynamics that project the newly computed full-space trial
positions and velocities onto the low-dimensional manifold, which effectively
updates the generalized coordinates and velocity.

Algorithm \ref{algo:projection_nonlinear} presents an approach that performs a
least-squares projection of the symplectic Euler updated position and velocity
onto the manifold and its tangent space, respectively. This associates with a
least-squares projection of the position and velocity, where the least-squares
problem is linear for the velocity, but nonlinear for the position; we solve
the latter using the Gauss--Newton method \citep{nocedal2006numerical} with
backtracking.  We refer to this approach as the position-velocity projection
scheme.

\begin{algorithm}[H]
    \SetAlgoLined
        \KwData{Full-space trial positions $\xTnpTrial$ and velocities ${\velTnpTrial}$ for $p\in\sampleSet$.}
        \KwResult{Generalized velocity $\genVels_{n+1}$ and generalized coordinates $\genCoords_{n+1}$.}

        $\genVels_{n+1}$ and $\genCoords_{n+1}$, which should satisfy the minimization problem
        \begin{align}
            \label{eqn:nonlinear_projection:vel}
            \genVels_{n+1} \in  \underset{\genVels\in\RR{\nred}}{\mathrm{arg\;min}}\ \sumParticlesSampleOnly\|
    \frac{\partial
            \deformationMapApprox}{\partial
            \genCoords}
            (\X^p;\genCoords_n) {\genVels} - {\velTnpTrial}\|_2^2.\\
            \label{eqn:nonlinear_projection:pos}
            \genCoords_{n+1} \in  \underset{\genCoords\in\RR{\nred}}{\mathrm{arg\;min}}\ \sumParticlesSampleOnly\|
            \deformationMapApprox
            (\X^p;\genCoords) - \xTnpTrial\|_2^2.
        \end{align}

    \caption{Reduced-space dynamics via position-velocity projection}
    \label{algo:projection_nonlinear}
\end{algorithm}

To reduce computational costs, we can linearize the nonlinear solve;
\Cref{sec:linearization} provides the derivation.\KTC{include reference to
the algorithm below}\PYC{done} The resulting approach is detailed in
Algorithm \ref{algo:projection_linear}. Since only velocity is involved in
the least-squares projection, we refer to this projection scheme as the
velocity-only projection scheme; in principle, it does not require the
full-space trial positions $\xTnpTrial$, $p\in\sampleSet$.

\begin{algorithm}[H]
    \SetAlgoLined
        \KwData{Full-space trial velocities ${\velTnpTrial}$ for $p\in\sampleSet$.}
        \KwResult{Generalized velocity $\genVels_{n+1}$ and generalized coordinates $\genCoords_{n+1}$.}
        $
    \genCoords_{n+1} = \genCoords_n + \timestepn\genVels_{n+1}
    $, where $\genVels_{n+1}$ satisfies the minimization problem
    \begin{align}\label{eqn:linear_projection}
        \genVels_{n+1} \in  \underset{\genVels\in\RR{\nred}}{\mathrm{arg\;min}}\ \sumParticlesSampleOnly\|
\frac{\partial
        \deformationMapApprox}{\partial
        \genCoords}
        (\X^p;\genCoords_n) {\genVels} - {\velTnpTrial}\|_2^2.
    \end{align}

    \caption{Reduced-space dynamics via velocity-only projection}
    \label{algo:projection_linear}
\end{algorithm}

\section{Manifold-parameterization construction via implicit neural
representation}
\label{sec:train}
\label{sec:manifold-construct}

\KTC{PC: Can you take a rev through this and emphasize the implicit neural
representation angle, also connecting to some prior work in the area?}\PYC{done.}
\begin{figure}[H]
  \centering
  \includegraphics[width=\textwidth]{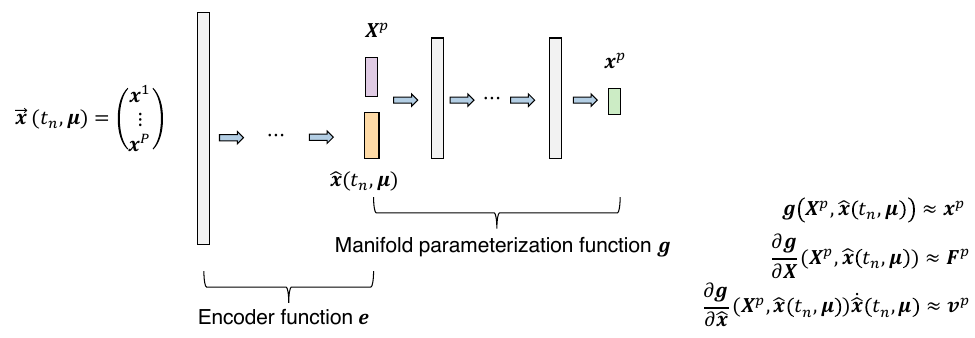}
  \caption{\rev{The manifold parameterization function $\vg$ is constructed via a
    multilayer perceptron neural network.} When we use a continuously differentiable
    activation function, we can also compute the approximated deformation gradient
    and the approximated velocity everywhere. An encoder network $\enc$ is used for
    generating $\genCoords$ from the simulation snapshot. \MMC{Please fix fonts
    - text and pictures ideally have the same font. There are tools out there
    that take a picture and create two layers of latex text and a layer of image
    only or any other tool.}\PYC{@MC see discussion earlier. inconsistent text and figures in the final? https://www.sciencedirect.com/science/article/pii/S2352431620301577} \KTC{I don't understand many things about the
    equations here. Why not simply make these equivalent to the statements in
    Section 3.1, which fleshed these out precisely? Also, why is there a finite
    difference approximation used for the last equation? Just take the time
    derivative of the generalized coordinates.}\PYC{1. sec 3.1 is before discretization; here we want to include discretization; 2. using approximation ($~=$) indicates training is an approximation; 3. finite difference removed.}}
  \label{img:network}
\end{figure}

In principle, the manifold-parameterization function
$\deformationMapApprox:\domainRef\times \RR{\nred}\rightarrow \RR{d}$ could be
constructed in various ways. In this work, we employ a fully-connected
  deep-learning architecture (i.e.,
a multilayer perceptron) for this
purpose (\Cref{img:network}), which yields an implicit neural representation
of the deformation map.\KTC{PC: can you recap a little bit here why this
architecture has been useful in the recent past?}\PYC{done.} Recently in the the computer vision community, this particular network structure has been shown to successfully model the signed distance fields of various geometries \citep{park2019deepsdf} and the radiance fields of different viewing directions \citep{mildenhall2020nerf}. We adopt ELU activations to
ensure continuous differentiability of the deformation-map approximation,
which is needed to compute the deformation gradient and velocity as expressed in
Eq.~\eqref{eq:defGrad} and Eq.~\eqref{eq:timederOne}, respectively.\MMC{ would expand here connecting to deformation
gradient and time derivative}\PYC{done}\KTC{We can skip this next section on the
inputs/outputs to $\vg$ as this was already covered back in Section 3.1. I
don't see any point in repeating the basic domain/codomain of this
function.}\PYC{I think it's still good to recap. added reference to sec 3.1} As discussed in \Cref{sec:mpkinematics}, the inputs to $\vg$ correspond to the undeformed position of
a material point $\vX^p\in\domainRef$ and the generalized coordinates $\genCoords(t,\params)\in\RR{\nred}$, the
latter of which is shared among all material
points. The output of $\vg$ is
the approximated deformed position of the material point $\vX^p$. Thanks to
the network's continuous differentiability, we also obtain the approximated
deformation gradient and approximated velocity via backpropagation. \KTC{do we use backprop or
forward propagation for these derivatives? We should say which here}\PYC{done.} 
\KTC{Removed everything related to our goals wrt training as those are
described in 4.2 below; this is just the architecture}

\subsection{Encoder}
\label{sec:encoder}
To train the manifold-parameterization function $\deformationMapApprox$, we
also need to define the value of the generalized coordinates $\genCoords$ at each time step $t_n$ for each
training parameter instance $\params\in\paramDomainTrain$. We do so implicitly by introducing an encoder
network $\enc$, which we train along with $\deformationMapApprox$. $\overrightarrow{\vx}(t_n,\params)$ is the input to $\enc$,
which is defined by concatenating the deformed positions of all the material
points.\KTC{Clarify that the encoder parameters are also exposed to training.
Also perhaps mention that is is not essential that \textit{all} material point
positions are included in the input; a subset could be selected in purpose to
keep training tractable with highly resolved training simulations.}\PYC{done.} Such an input encourages the injectivity of $\deformationMapApprox$
with respect to $\genCoords$ since there exists a unique $\genCoords$ that
corresponds to a simulation state, as defined by all the positions of the
material points.\KTC{Disagree with the previous statement. Nothing
in the framework prevents many instances of $\overrightarrow{\vx}$ mapping to the same
generalized coordinates, which would preclude injectivity.}\PYC{@KC, discuss; if different $\overrightarrow{\vx}$ maps to the same xhat via encoder, then there is no way the decoder maps to different $\overrightarrow{\vx}$ since the input xhat is the same. perhaps, you meant something else?} This input is particularly suitable for history-independent
problems, e.g., elasticity. For history-dependent problems, history-dependent
variables can also be concatenated to the input to the encoder function to
define a simulation state uniquely.\KTC{Again, don't understand the argument
in the previous section. I suggest deleting, as we haven't handled this case
anyway.}\PYC{disagree; i have had ppl asking me why the input to encoder involves only position but not velocity. this is because our problem is not rate-dependent. otherwise, need to include more information to differentiate simulation state.} In practice, it is not essential that \textit{all} material point
positions are included in the input; a subset could be selected in purpose to
keep training tractable with highly resolved training simulations. In addition, the encoder structure also
encourages a spatially and temporally coherent representation of the
simulation state where contiguous generalized coordinates\KTC{`nearby $\genCoords$s' is too colloquial}\PYC{contiguous} correspond to nearby simulation
states \citep{bengio2013representation}.
\begin{remark}
    \KTC{Improve the precision here.}\PYC{done.} While we utilize a
		neural-network-based encoder to associate each training sample with a
		value for the generalized coordinates, this association can also be
		made in other ways. For
    example, the generalized coordinates for each training sample can be
		defined by explicit, manual choice or by
    exposing the generalized coordinates for each training sample to the
		optimization algorithm as training variables along with the network weights. In practice, we found these approaches
    challenging to scale to problems involving a large number of training
		samples, as the number of optimization variables with this approach scales
		linearly with the number of training samples. \KTC{`number of $\genCoords$s' is far too colloquial. Do you mean
    `the dimensionality of $\genCoords$?}\PYC{edited}Furthermore, it is inconvenient to enforce spatial
    and temporal coherence with these techniques. \KTC{again, I don't understand
    how the approach above 'enforces injectivity', so I'd remove this}\PYC{done.}
\end{remark}

\subsection{Loss function}
    We compute the neural-network weights $\theta_{g}^\star$ and
    $\theta_{e}^\star$ of the functions $\deformationMapApprox$ and $\enc$ as
    the (approximate) solutions to the minimization problem
    \KTC{Check the limits of summation, you have time going beyond the final
    number of time steps for example}\PYC{no problem here; zeroth time is also used for training.}
   \begin{align}
   \begin{split}
   \underset{\theta_g,\theta_e}{\mathrm{minimize}}\
     &\sum_{n=0,\ldots,\nTimesteps,\;p=1,\ldots,\nParticles,\;\params\in\paramDomainTrain}
   (\|\deformationMapApprox_{\theta_g}(\X^p;\genCoordsArgsParam{t_n}{\params}) -
   \deformationMap(\X^p;t_n,\params) \|_2^2\\
     &+\lambda_F
 \|\nabla\deformationMapApprox_{\theta_g}(\X^p;\genCoordsArgsParam{t_n}{\params}) -
   \nabla\deformationMap(\X^p;t_n,\params) \|_F^2\\
    &+\lambda_v
    \|\frac{\partial\deformationMapApprox_{\theta_g}}{\partial\genCoords}(\X^p;\genCoordsArgsParam{t_n}{\params})
    \frac{\genCoordsArgsParam{t_{n+1}}{\params}-\genCoordsArgsParam{t_n}{\params}}{\Delta t}-
   \dot\deformationMap(\X^p;t_n,\params) \|_2^2
   )
   \end{split}
   \end{align}
     where
     $\genCoordsArgsParam{t_n}{\params}\defeq\enc_{\theta_e}(\overrightarrow{\vx}(t_n,
     \params))$,
     $\paramDomainTrain\subseteq\paramDomain$ denotes the parameter instances for training, at
   which the full-order model has been solved and solutions are available, and
   $\lambda_F,\lambda_v\in\RR{}_+$ denote penalty parameters for the
   deformation gradient and velocity, respectively. 
   
     The deformation-gradient penalty $\lambda_F$ serves to enable the network
     to generate a manifold that can accurately represent the spatial gradients (Eq.~\eqref{eq:defGrad}), which is
     essential for both Lagrangian-quadrature kinematics (Algorithm
     \ref{algo:evolution-qmpm}) and Eulerian-quadrature kinematics (Algorithm
     \ref{algo:evolution-qfem}). The velocity penalty $\lambda_v$ serves to
     enable the network to generate a manifold whose tangent space can
     accurately capture the velocity (Eq.~\eqref{eq:timederOne}). This concept could
     be applied to higher-order spatiotemporal derivatives if desired. The practical choices of $\lambda_F$ and $\lambda_v$ are detailed in the result section (\Cref{sec:experiments}).  \MMC{If I were to implement this,
     how would I choose these values - please remark that this is discussed in a
     later section.}\PYC{done.}

     Note that the
     $\frac{\genCoordsArgs{t_{n+1}}{\params}-\genCoordsArgs{t_n}{\params}}{\Delta
     t}$ term is a finite difference approximation of the generalized velocity. Such an approximation mitigates the truncation error incurred by the linearization underpinning the
		 velocity-only projection scheme for reduced-space dynamics (\Cref{sec:linearization}). \KTC{`effectively
		 guarantees' is too strong without backing analysis. Perhaps `mitigates
		 truncation error incurred by the linearization underpinning the
		 velocity-only projection scheme for reduced-space dynamics'.}

\section{Numerical experiments}
\label{sec:experiments}
We demonstrate the robustness of the proposed reduced-order approach on
several large-deformation nonlinear elasticity problems with complex geometry.\KTC{Added `with complex geometry'} The particular
constitutive law we adopt is the fixed corotated hyperelastic energy by
\citet{stomakhin2012energetically}; in principle, any hyperelastic model is
compatible
with the proposed approach without modification. We employ
the open-source explicit MPM implementation
by \citet{wang2020hierarchical} to define our baseline full-order model.
Both the full-order and 
reduced-order models run on 12 threads on a 2.30GHz Intel Xeon E5-2686 v4 CPU.
In addition, the neural network portion of the reduced-order model
pipeline---which comprises evaluation, inversion, and differentiation of the
deformation-map approximation\KTC{Peter - verify this parenthetical}\PYC{good}---is
implemented using the LibTorch library \citep{paszke2019pytorch} and runs on a
single NVIDIA Tesla V100 GPU.

\begin{table}
  \centering
  \begin{tabularx}{\textwidth}{C}
      Encoder network $\enc$ \\
  \end{tabularx}
  \begin{tabularx}{\textwidth}{C}
      \toprule
      1D convolution layers\\
  \end{tabularx}
  \begin{tabularx}{\textwidth}{CCC}
      \midrule
          Layer & Kernel size & Stride size\\
      \midrule
      $1$ & $6$ & $2$\\
      $\ldots$ & $6$ & $2$\\
      $n_{conv}$ & $6$ & $2$\\
      \midrule
  \end{tabularx}
  \begin{tabularx}{\textwidth}{C}
      Fully-connected layers\\
  \end{tabularx}
  \begin{tabularx}{\textwidth}{CCC}
  \midrule
      Layer & Input dimension & Output dimension\\
  \midrule
  $n_{conv}+1$ & $d_{conv}$ & $32$\\
  $n_{conv}+2$ & $32$ & $\nred$\\
  \bottomrule
  \end{tabularx}

  \begin{tabularx}{\textwidth}{C}
      \\
  \end{tabularx}

  \begin{tabularx}{\textwidth}{C}
      Manifold-parameterization function $\deformationMapApprox$ \\
  \end{tabularx}
  \begin{tabularx}{\textwidth}{C}
      \toprule
      Fully-connected layers\\
  \end{tabularx}
  \begin{tabularx}{\textwidth}{CCC}
  \midrule
      Layer & Input dimension & Output dimension\\
  \midrule
  1 & $d + \nred$ & $30$\\
  2 & $30$ & $30$\\
  3 & $30$ & $30$\\
  4 & $30$ & $30$\\
  5 & $30$ & $d$\\
  \bottomrule
  \end{tabularx}
  \caption{For the manifold-parameterization function $\deformationMapApprox$,
  we adopt a lightweight implicit neural representation network by using 5 fully connected hidden layers,
  each of size 30, where $d \in \{2,3\}$ and $\nred$ denotes the
	reduced-order-model dimension. For the encoder function $\enc$, since the
  concatenated input vector can be arbitrarily large depending on the number
	of material points $\nParticles$ in the full-order-model simulation, we avoid extensive usages of fully connected layers.
  Instead, several 1D convolution layers with a kernel size of 6 and a stride
  size of 2 are used to reduce the dimension of the input vector down to
  $d_{conv}$, which is as low as possible but no smaller than 32. After that,
  a fully connected layer transforms the previous layer into a vector of size
  32. Another fully connected layer then transforms the previous layer into
  a vector of the size $\nred$, the dimension of the generalized coordinate.
  \KTC{Way too detailed for the main body, should be in the experiments
  section}}\PYC{done.} \label{tbl:network_details}
\end{table}

\Cref{tbl:network_details} lists the detailed network structure of the manifold-parameterization function $\deformationMapApprox$ and the encoder function $\enc$. We adopt this network structure for all experiments presented in this work. The rest of the training details are listed in \Cref{sec:training_details}.

For hyper-reduction (\Cref{sec:hyper-reduction}), we find sampling at least
5 material points from each kinematic boundary generates stable
reduced-order dynamics.\KTC{I don't like these qualitative
statements that aren't backed by evidence. What is meant by `provides a good
approximation of the BC'? Perhaps something like `We find sampling at least
5 material points from each kinematic boundary generates stable
reduced-order dynamics.'}\PYC{done.} For the Eulerian quadrature point scheme (\Cref{sec:quadratures}), we use $\ell =2$, i.e., 2 quadrature points per cell per dimension. For the position-velocity projection scheme (\Cref{sec:projection}), we use a simple linear interpolation of the previous generalized coordinates as the initial guess, $\genCoords^\text{guess} =
2\genCoords_{n}-\genCoords(t_{n-1}$), and the solver typically converges in
2--3 iterations.

All reported errors in the following sections correspond to the accumulated position errors \rev{(\%)} of the test
simulations executed at parameter instances not included in the set employed
for training, i.e.,
\begin{align*}
  \text{position error \rev{(\%)}} = 
   \rev{100*}\frac{
     \sqrt{\sum\limits_{n=0,\ldots,\nTimesteps,\;p=1,\ldots,\nParticles,\;\params\in\paramDomainTest}
     \|\deformationMapApprox(\X^p;\genCoordsArgs{t_n}{\params}) -
   \deformationMap(\X^p;t_n,\params) \|_2^2
   }}
   {
     \sqrt{\sum\limits_{n=0,\ldots,\nTimesteps,\;p=1,\ldots,\nParticles,\;\params\in\paramDomainTest}
    \|\deformationMap(\X^p;t_n,\params) \|_2^2}
   }.
\end{align*}
Here, $\paramDomainTest\subseteq\paramDomain$ with
$\paramDomainTest\cap\paramDomainTrain=\emptyset$ denotes the set of test parameter
instances. Note that this approach to error estimation is not practical for
real applications, as it requires executing the full-order model and
evaluating the discrepancy between the full-order and reduced-order solutions
for all material points at every time instance. Future work will pursue
applying approaches that generate low-cost, statistically validated models of
the ROM error \cite{freno2019machine,parish2020time}.

\begin{table}
  \centering

  \resizebox{\columnwidth}{!}{
    \begin{tabular}{lllllllllll}
      \toprule
      Experiment & Geometry & Young's & Poisson & \# of & Grid cell & Particles per cell & Time step & Time steps & \# of training  & \# of testing\\
      & & modulus & ratio & particles & width &  per dimension & size  & per simulation & simulations & \ simulations\\
      \midrule
      \Cref{sec:gravity} & Cylinder  & $12500$ Pa & $0.3$ & $1,368$ & $0.04$ cm & 2 & $\frac{1}{144}$ s & 30 & 24 & 6\\
      \Cref{sec:torsion-tension} & Cuboid & $12500$ Pa & $0.3$ & $1,757$ & $0.04$ cm & 2 & $\frac{1}{144}$ s & 30 & 29 & 7\\
      \Cref{sec:poke-and-recover} & Cylinder  & $12500$ Pa & $0.3$ & $1,368$ & $0.04$ cm & 2 & $\frac{1}{144}$ s & 432 & 12 & 24\\
      \Cref{sec:application} & Tower & $80000$ Pa & $0.2$ & $3,076,115$ & $0.48$ cm & 3 & $\frac{1}{24}$ s & 80 & 16 & 4\\
      \bottomrule
    \end{tabular}
  }

    \caption{Material properties and discretization parameters. Material
    points are initially positioned through the Poisson disk sampling approach
    \citep{bridson2007fast} by sampling a fixed number of particles\KTC{not
    particles, material points; fix everywhere}\PYC{@MC, disagree, particle is the standard terminology, consistent with the literature.} per grid cell per dimension \citep{jiang2016material}. The time step size is computed from the stability analysis based on the speed of the elastic wave and the element characteristic length scale \citep{fang2018temporally}.} \label{tbl:example_parameters} 
\end{table}

\subsection{Gravity}
\label{sec:gravity}
\begin{figure}
   \centering
   \includegraphics[width=0.25\linewidth]{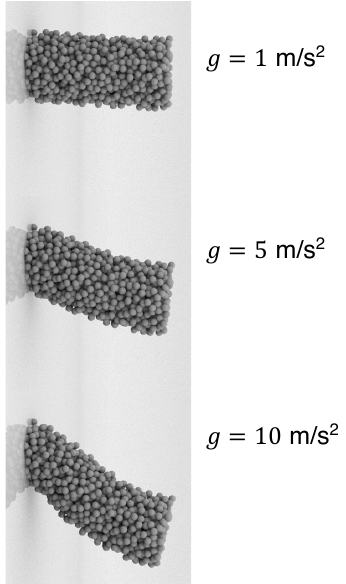}
   \caption{The reduced-order simulation handles a wide range of gravity
   values. All snapshots are taken at the 30th time step ($t=0.208$ s). \KTC{at what times are these snapshots taken?}\PYC{done.}}
   \label{gravity}
 \end{figure}

We conduct our first set of numerical experiments on an elastic cylinder with a radius of
$1$ cm and a height of $4$ cm. Its material and discretization parameters\KTC{replaced `numerical properties' with `discretization parameters' as I've
never heard the former term}\PYC{changed throughout} are listed in \Cref{tbl:example_parameters}. The elastic cylinder is attached to a
vertical wall on one side and deforms under the influence of
a downward gravity (\Cref{gravity}).

We consider the parameterized problem $\mub=g \in \paramDomain\subseteq\RRPlus$, where $g$ denotes the
magnitude of the gravitational force. We generate training and testing data
via uniform sampling
of $30$ points in the interval $g\in[1, 10]$ m/s$^2$. For each value of $g$,
we execute a simulation of
$30$ time steps. Therefore, a total of $930$ simulation snapshots are
generated, including the initial conditions. We then randomly split the
dataset of $30$ simulations into an offline training dataset of 24 full simulations
and an online testing dataset of 6 full simulations. \KTC{Is it true that we are
only varying one parameter in this example, i.e., $\nparams=1$? We should
clearly state this}\PYC{done.}

\subsubsection{The effect of gradient penalties}
\label{sec:effect_grad_pen}
\begin{figure}
   \centering
   \includegraphics[width=\textwidth]{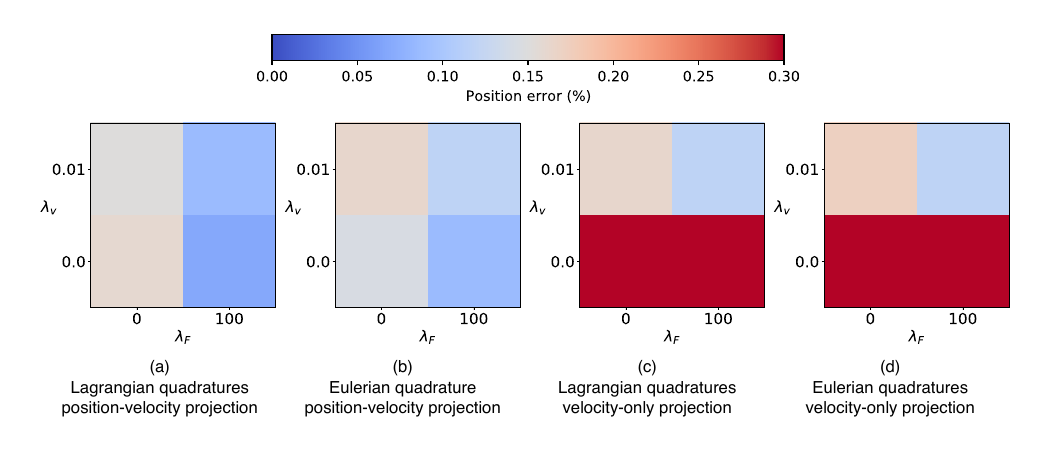}
   \caption{The effect of gradient penalties during training. The temporal penalty term $\lambda_v$ improves the accuracy of the
   velocity-only projection schemes (c and d) but not the position-velocity projection schemes (a and b). The
   spatial penalty term $\lambda_F$ improves both projection schemes. The
   Lagrangian (a and c) and the Eulerian (b and d) quadrature approaches yield
   similar results. Note that the experiment setup considers the parameterized
	 problem $\mub=g \in \paramDomain\subseteq\RRPlus$, where $g$ denotes the
   magnitude of the gravitational force. No hyper-reduction is conducted. \KTC{Mention the dimension of the ROM and other important
   parameters (e.g., hyper-redution) in the caption}\PYC{done.} \KTC{Just checking, is
   the red result 0.3\% or 30\%? It currently reads as 0.3\%, which really
   isn't too bad}\PYC{see intro.}}
   \label{img:gravity_penalty}
 \end{figure}

 In \Cref{img:gravity_penalty}, we study the influence of \emph{offline}
 training parameters on the accuracy of the \emph{online} reduced-order
 simulation. The dimension of the generalized
 coordinates is fixed to be 7; similar trends are observed for other
 generalized-coordinates dimensions. After training, we conduct reduced-order
 simulations using the four different combinations proposed in
 \Cref{sec:dynamics} and Algorithm \ref{algo:reduced_order_dynamics}. To
 isolate the source of error, we do not conduct hyper-reduction here.

 Training with a nonzero value of the velocity-penalty parameter $\lambda_v$ significantly improves the accuracy of
 the velocity-only projection technique of Algorithm \ref{algo:projection_linear}
 (\Cref{img:gravity_penalty} c and d). By contrast, the position-velocity projection
 scheme of Algorithm \ref{algo:projection_nonlinear}
 (\Cref{img:gravity_penalty} a and b) is less sensitive to the choice of
 $\lambda_v$. Such observations hold for both the Lagrangian quadratures and
 the Eulerian quadratures. This numerical result aligns well with the
 theoretical analysis of linearization (\Cref{sec:linearization}), because the velocity-only projection technique relies primarily on the accuracy of the velocity projection.\KTC{Isn't it also intuitive, as `velocity-only projection' (name to be fixed) relies primarily on accurate velocity projection followed by integration of the generalized coordinates?}\PYC{yes,add a quick explanation} Furthermore, training with a nonzero value of the deformation-gradient penalty $\lambda_F$
 improves all four algorithm combinations (\Cref{img:gravity_penalty} a, b, c,
 and d), likely due to the fact that this penalty encourages better
 deformation-gradient approximations that are essential for defining the
 quadrature-point kinematics.
 \KTC{any idea why $\lambda_v$ hurts accuracy for position-velocity projection with
 $\lambda_F$ fixed to 100?}\PYC{maybe lambdaF is not as accurate? I would not focus on the diff there as it is still comparable accuracy, in comparision to the really red result.}

 \subsubsection{The effect of the generalized coordinates dimension}
 \label{sec:effect_gencoorddim}
 \begin{figure}
  \centering
  \includegraphics[width=0.6\textwidth]{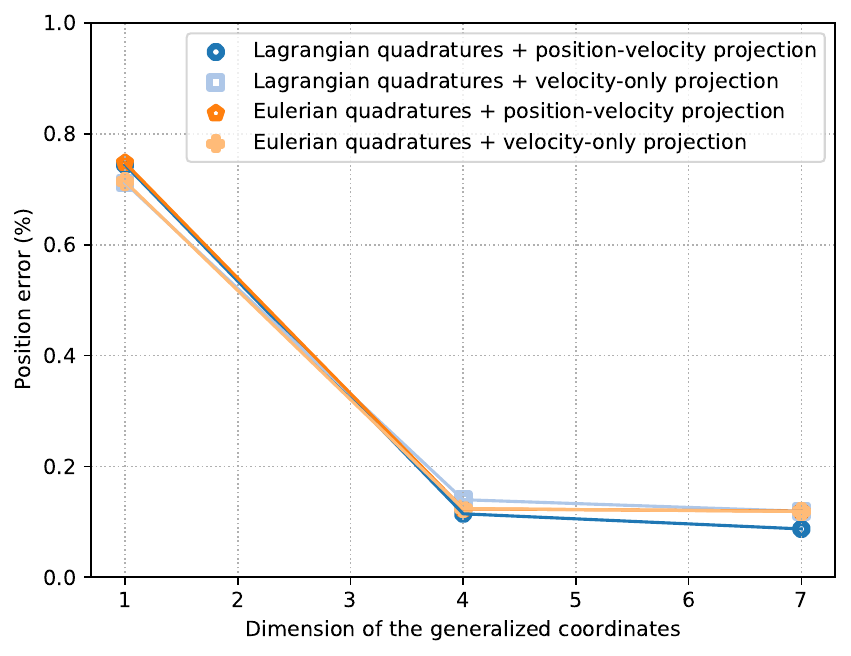}
  \caption{Increasing the dimension of the generalized coordinates improves the
   accuracy due to larger manifold dimensions. \KTC{Don't like `larger learning
   capacity'. I'd attribute this to `larger manifold dimension'}\PYC{done.}}
  \label{img:gravity:gencoord_dim}
\end{figure}

\Cref{img:gravity:gencoord_dim} demonstrates the effect of the generalized coordinates dimension $\nred$, which is a key hyperparameter of the network structure (\Cref{sec:manifold-construct}). We train networks with different generalized coordinates dimensions while fixing $\lambda_v$ to be 0.01 and $\lambda_F$ to be 100. Afterward, the trained networks are tested for reduced-order simulations. In order to isolate the source of error, hyper-reduction is not applied. \Cref{img:gravity:gencoord_dim} shows that the four different algorithm combinations from Algorithm \ref{algo:reduced_order_dynamics} demonstrate the same trend: increasing the generalized coordinates dimension improves the simulation accuracy because the network has a larger manifold dimension.\KTC{Don't like `larger learning
   capacity'. I'd attribute this to `larger manifold dimension'.}\PYC{done.}

\subsubsection{The effect of hyper-reduction}
\label{sec:effect_hyperreduction}
\begin{figure}[H]
  \centering
  \includegraphics[width=0.6\textwidth]{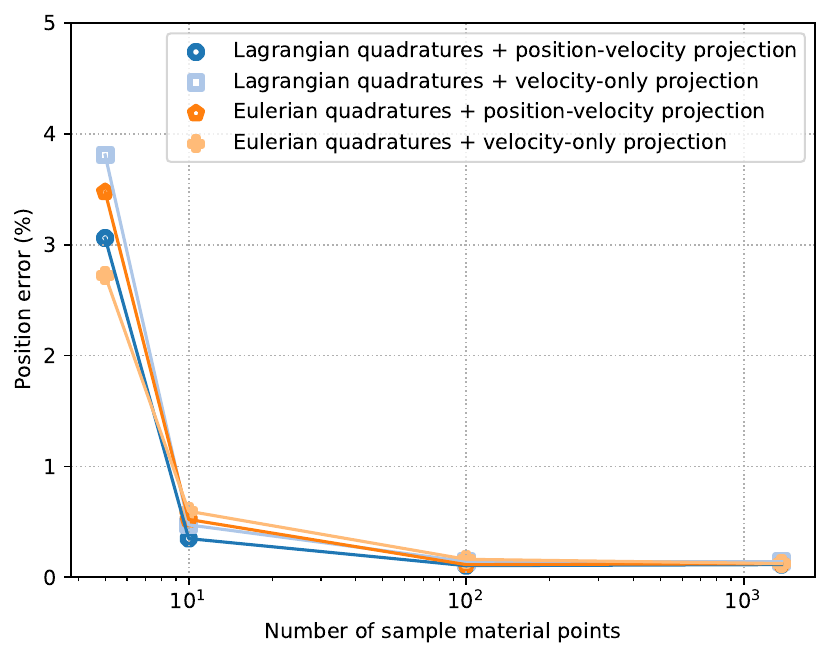}
  \caption{The effectiveness of hyper-reduction is evidenced by the fact that using 10 sample material
  points yields less than 1\% error, and using just 100 points yields the same accuracy as no hyper-reduction, where all 1,368 material points are used for projection.}
  \label{img:gravity:hyper-reduction}
\end{figure}
\KTC{Note that everywhere else in the paper, we used the term `sample material
points'; suddenly, this setion uses the term `projection points'. We have to be
completely consistent}\PYC{done.}
\Cref{img:gravity:hyper-reduction} reports the influence that the number of
sample material
points has
on the accuracy of the reduced-order simulations. After the offline
training with a setup of $\lambda_v=0.01$, $\lambda_F=100$, $\nred=4$, we
conduct online, reduced-order simulations with various numbers of sample material points. Notably, with just 10 sample material
points, all four quadrature
and projection combinations yield an error of less than 1\%. In addition,
using just 100 points delivers the same level of accuracy as no
hyper-reduction, i.e., all 1,368 points' dynamics are computed
(\Cref{sec:evolution}) and used for projection onto the generalized
coordinates (\Cref{sec:projection}).

\subsubsection{Hyperparameter summary}
\label{sec:hyper-sum}
\begin{figure}
  \centering
  \includegraphics[width=\textwidth]{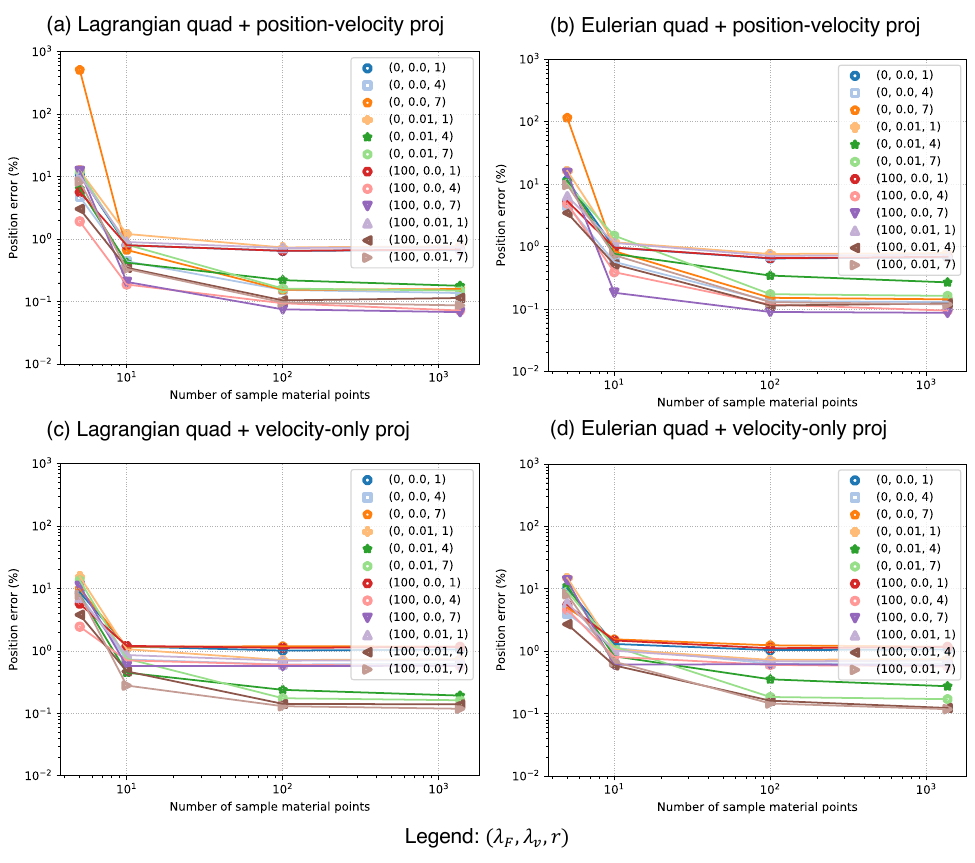}
  \caption{Hyperparameter summary. Training with positive spatial and temporal
	gradient penalties yields the best results. The size of the generalized
	coordinates should be larger than 1 in order to attain the best
	accuracy. The use of at least 10 sample material points leads to high
	projection accuracy.}
  \label{img:gravity_summary}
\end{figure}

To summarize all the offline and online hyperparameter options, we plot all the choices together (\Cref{img:gravity_summary}). \Cref{sec:effect_grad_pen}, \Cref{sec:effect_gencoorddim}, and \Cref{sec:effect_hyperreduction} each presents a ``slice'' of the hyperparameter study in \Cref{img:gravity_summary} in order to articulate the effect of a particular parameter.

As shown in \Cref{img:gravity_summary}, independent of quadrature and
projection combinations, training with a generalized-coordinate dimension of
$\nred=1$\KTC{PC: Always avoid referring to symbols directly in writing, as
the averagge writer won't be able to keep all their meanings in their head.
Rather, refer to the term itself and then use the symbol, e.g., `the
generalized-coordinater dimension $\nred$ is 5'. Recommend doing one sweep
on the paper to identify/fix these instances; will be a good long-term writing
habit to acquire.}\PYC{will do} always yields a worse result, e.g., $(100,0.01,1)$ vs. $(100,0.01,4)$, $(100,0.01,1)$ vs. $(100,0.01,7)$. Therefore, the default training strategy should use $\nred>1$. Unlike the position-velocity projection scheme (\Cref{img:gravity_summary} a and b), the velocity-only projection scheme (\Cref{img:gravity_summary} c and d) also consistently yields better result when training with a positive temporal gradient penalty $\lambda_v>0$, e.g. $(0,0.0,4)$ vs. $(0,0.01,4)$, $(0,0.0,7)$ vs. $(0,0.01,7)$. Therefore, the default training strategy should use a positive temporal gradient penalty, especially when using velocity-only projection. The spatial gradient penalty also improves simulation accuracy, as discussed in \Cref{sec:effect_grad_pen}. Therefore, the default training strategy should also include the spatial gradient penalty term, though its significance is lesser than the other terms (\Cref{img:gravity_summary}). With the training strategy mentioned earlier, all four reduced-order schemes achieve less than 1\% error with just 10-100 sample material
points. To obtain meaningful results, projection with just one point should be avoided as it can cause an error over $100\%$.

Since this section focuses on small-scale experiments that serve to support an
extensive parameter study; as such, the opportunity for wall-time speedup of
the reduced-order method over the full-order method is diminished. Experiments
in \Cref{sec:application} will report wall-time speedups for higher-dimension
problems. \KTC{never use a double semicolon. recommend `diminished.
Experiments...'}\PYC{done, just one semicolon now.}
 
\subsection{Torsion and tension}
\label{sec:torsion-tension}
\begin{figure}
   \centering
   \includegraphics[width=0.4\linewidth]{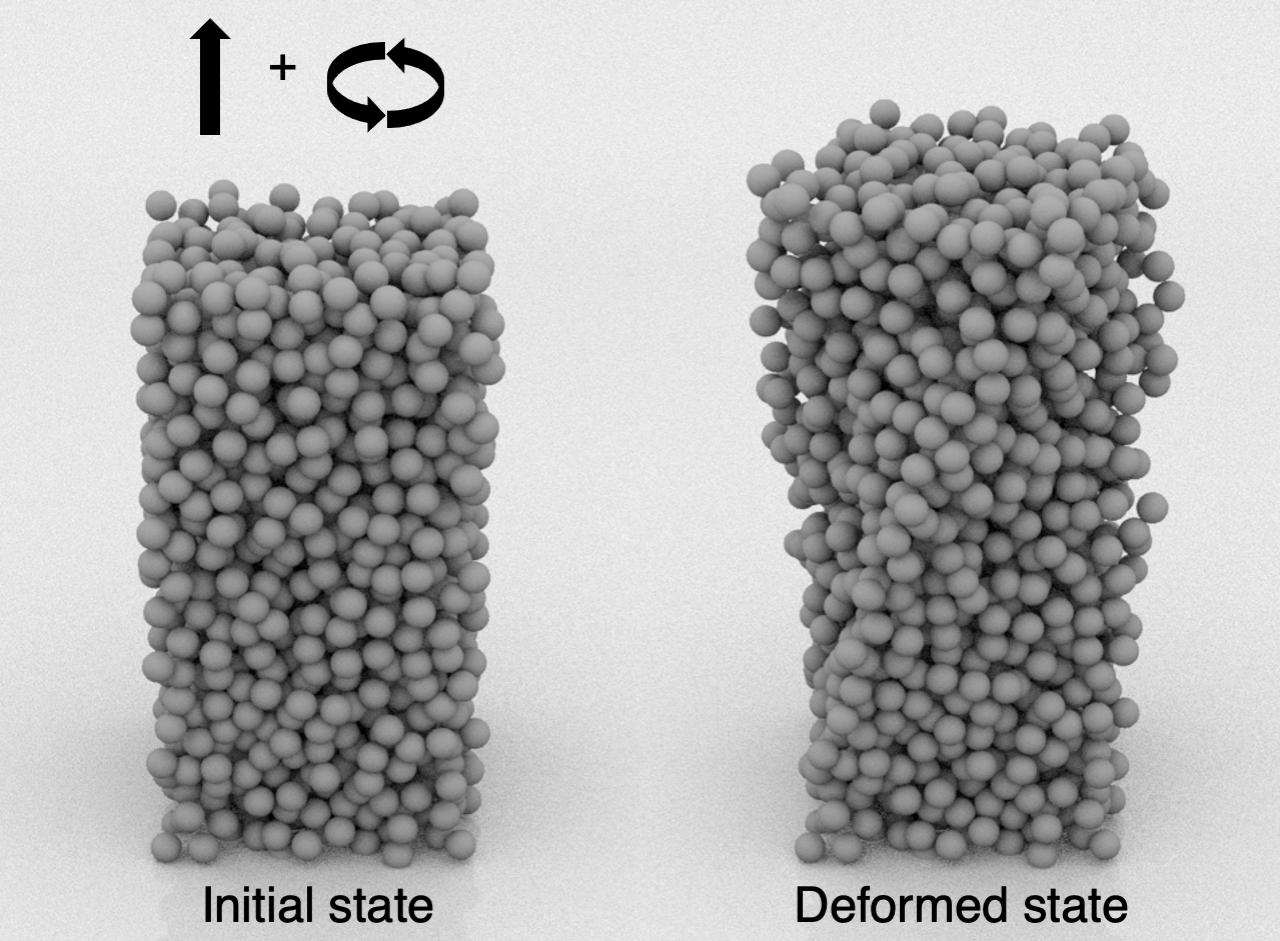}
   \caption{The object undergoes tension and torsion at the same time.}
   \label{img:torsion-tension}
 \end{figure}
We apply tension and torsion to an elastic, rectangular
cuboid (\Cref{img:torsion-tension}). The dimensions of the cuboid are $1$ cm,
$1$ cm, and $4$ cm. Its material and discretization properties are listed in \Cref{tbl:example_parameters}.

We consider the parameterized problem $\mub=(v, \omega) \in \paramDomain\subseteq\RR{2}$, where $v$ denotes the translational velocity and $\omega$ denotes the rotational velocity. We generate simulation data by varying the translational
velocity ($v\in[0, 0.6]$ m/s) and the rotational velocity ($\omega\in[0, 2]$
rad/s).\KTC{To be completely clear, please recap the total number of varying
parameters in this example}\PYC{done.} A total number of $36$ simulations are generated
via full factorial
sampling of the translation and the rotational velocities with six evenly spaced
samples in each dimension\rev{, i.e., $v\in\{0, 0.12, 0.24, 0.36, 0.48, 0.6\}$ and $\omega\in\{0, 0.4, 0.8, 0.12, 0.16, 2\}$}. Each simulation consists of $30$ time steps.
Therefore, we generate a total of $1,116$ simulation snapshots, including the initial
conditions.\KTC{`are generated' is passive voice; `We
generate...'}\PYC{done.} Afterward, we randomly assign $29$ full simulations for
training and $7$ for testing. \rev{Specifically, the testing parameters are $(v, \omega) \in \{(0.6, 2), (0.6, 0.4), (0.24, 0.8), (0.36, 1.6), (0.12, 0.8), (0.24, 1.2), (0.6, 1.6)\}$ and the rest are used for training.}

\begin{table}
  \centering
    \begin{tabular}{lrr}
      \toprule
      Scheme & Sample material & Sample material\\
      & points count: 50 & points count: 1,757 (all)\\
      \midrule
      Lagrangian quad + position-velocity proj & $0.28\%$ & $0.22\%$\\
      Lagrangian quad + velocity-only proj & $0.39\%$ & $0.29\%$\\
      Eulerian quad + position-velocity proj & $0.33\%$ & $0.24\%$\\
      Eulerian quad + velocity-only proj & $0.34\%$ & $0.27\%$\\
      \bottomrule
      \end{tabular}
      \caption{Torsion and tension: errors of the reduced-order simulations on
      the testing dataset.} \label{tbl:roten}
  \end{table}

An approximated deformation map network is trained with $\lambda_F = 100, \lambda_v=0.01, \nred=7$ (cf. \Cref{sec:hyper-sum}). We then conduct reduced-order simulations using this network. \Cref{tbl:roten} reports the testing errors of the reduced-order simulations using different quadrature and projection combinations with and without hyper-reduction, demonstrating the effectiveness of the reduced-order simulation in modeling tension and torsion.

\subsubsection{Zero-shot super-resolution}

\begin{figure}
  \centering
  \includegraphics[width=\textwidth]{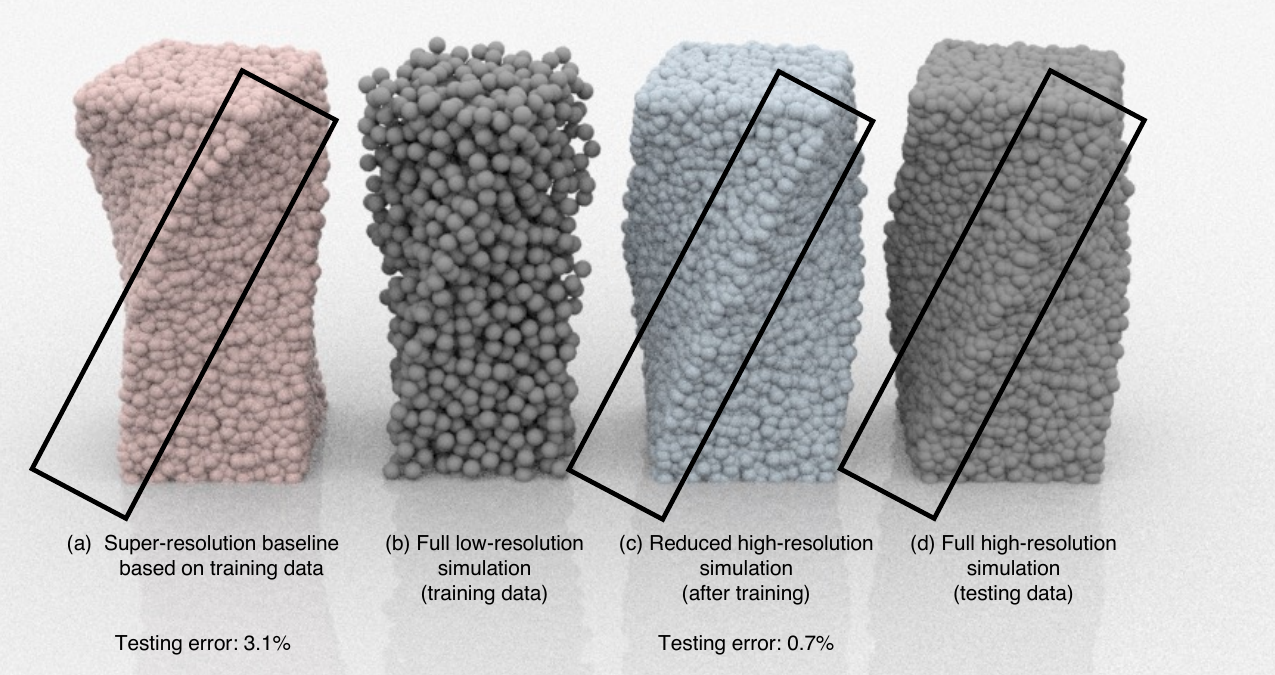}
  \caption{The approximated deformation map is trained with low-resolution
	simulations (b). We can then run high-resolution, reduced-order simulations
	(c) that agree well with high-resolution, full-order simulations (d). The
	low-resolution simulation has 1,757 material points
	(\Cref{tbl:example_parameters}). The high-resolution simulation increases
	the spatial resolution by two times in each dimension and has 13,900
	material points in its initial setup. The Poisson disk sampling approach
	(\Cref{tbl:example_parameters}) employed for generating initial material
	points has a random nature. None of the high-resolution material points
	shares identical initial positions with the low-resolution material points.
	In (a), we construct a super-resolution baseline\KTC{you should move the
	description of the super-resolution baseline to the main text; I missed it
	because many view the captions as somewhat optional reading.}\PYC{done.} by using the ``tracer particle technique'' \citep{stam1999stable,fu2017polynomial}, where we advect the high-resolution material points using the velocity field computed from the low-resolution simulation. In comparison, our reduced high-resolution simulation has a higher accuracy than the baseline, both visually and quantitatively, measured by the position error from the high-resolution simulation ground truth. Note that our model (c) produces the same straight boundary as the ground truth testing data (d), as highlighted in the region inside the black rectangle. By contrast, the baseline model (a) has an incorrect curved boundary.}
  \label{img:torsion-tension:super-res}
\end{figure}

\begin{table}
  \centering
    \begin{tabular}{lrr}
      \toprule
        Scheme & Position error\\
      \midrule
      Lagrangian quad + position-velocity proj & $0.49\%$\\
      Lagrangian quad + position-velocity proj (w/hyper-red) & $0.53\%$\\
      Lagrangian quad + velocity-only proj & $0.48\%$\\
      Lagrangian quad + velocity-only proj (w/hyper-red) & $0.63\%$\\
      Eulerian quad + position-velocity proj & $0.47\%$\\
      Eulerian quad + position-velocity proj (w/hyper-red) & $0.50\%$\\
      Eulerian quad + velocity-only proj & $0.46\%$\\
      Eulerian quad + velocity-only proj (w/hyper-red) & $0.51\%$\\
      \midrule
      Super-resolution baseline & $2.04\%$\\
      \bottomrule
      \end{tabular}
      \caption{Zero-shot super-resolution: errors of the reduced-order
      simulations on the high-resolution, torsion and tension testing dataset. Our framework outperforms the super-resolution baseline without and with
			hyperreduction (using 50 sample material points of the original 13,900
			high-resolution material points).\KTC{This is a great result, but needs
			to have a full description of the baseline in the text so that it can be
			better appreciated. Nice!}\PYC{done.}} \label{tbl:roten:zero:super-res}
  \end{table}

An advantage of training the deformation map instead of a \emph{finite} number
of material points is that we can easily adjust the resolution of the
reduced-order simulation. We can infer the dynamics of an \emph{infinite}
number of material points so long as they belong to the reference domain.
Consequently, even though the deformation map is trained on low-resolution
simulations (\Cref{img:torsion-tension:super-res}b), we can run
high-resolution reduced-order simulations
(\Cref{img:torsion-tension:super-res}c) by using a finer MPM grid. Since the high-resolution simulation is
never exposed to the training process, zero-shot super-resolution is achieved
\citep{li2021fourier}. We construct a super-resolution baseline by advecting
high-resolution material points on the velocity field computed by the low-resolution
simulation (a), employing the popular ``tracer particle technique'' \citep{stam1999stable,fu2017polynomial}. Both simulations are compared with the high-resolution ground
truth (\Cref{img:torsion-tension:super-res}d). \Cref{tbl:roten:zero:super-res}
lists the errors of the super-resolution simulations of all the reduced-order
schemes and the baseline, across the entire testing dataset. All reduced-order
methods outperform the super-resolution baseline significantly. \Cref{img:torsion-tension:super-res} demonstrates the visual superiority of our approach in comparison with the baseline approach.\KTC{What would
super-resolution with hyper-reduction look like? We'd need Eulerian
quadrature, right? Could we add to this example one hyper-reduction? Seems
likely we would beat the super-resolution baseline, right?}\PYC{Done. Lagrangian and Eulerian quadrature both work.}

\subsection{Poke-and-recover}
\label{sec:poke-and-recover}

\begin{figure}
  \centering
  \includegraphics[width=0.8\linewidth]{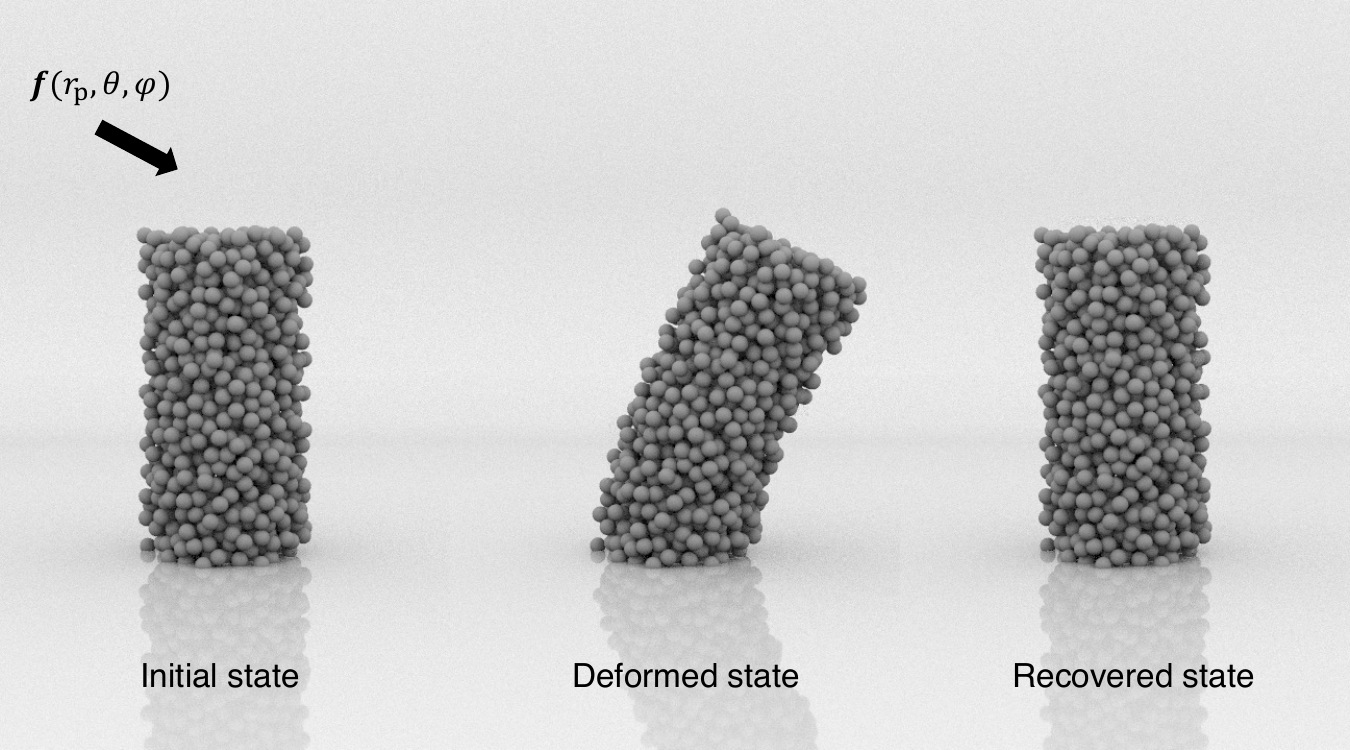}
  \caption{The material is poked at the top by different forces, resulting in different deformed states. The material then recovers to its initial state because of elasticity.}
  \label{img:poking-test}
\end{figure}

Poking is a frequent use case of the elasticity simulation where a force is applied in a particular direction at a small portion of the material and is released after a short period. The material then recovers to its undeformed state due to elasticity.

The elastic cylinder from \Cref{sec:gravity} is poked at the top
(\Cref{img:poking-test}). The poking force is characterized by the spherical
coordinate, where $\boldsymbol{f}(\radpok, \theta,
\phi)=(\radpok\sin\phi\cos\theta,\radpok\sin\phi\sin\theta,\radpok\cos\phi)$. The
corresponding poking location, at which the force is applied, is
$(-\radcyl\cos\theta, -\radcyl\sin\theta, h)$, where $\radcyl$ and $h$ are the radius and the
height of the cylinder, respectively. $\radpok$ is chosen such that the poked
location moves at a constant speed of 4.8 cm/s. $\phi$ is fixed to be
$\frac{1}{12}\pi$. The poking force is applied for 0.25 s before it is
released. After the force is released, the cylinder recovers to its initial
state due to elasticity. 

We consider the parameterized problem $\mub=\theta \in
\paramDomain\subseteq[0,2\pi)$. We generate simulation data via uniform
sampling of 
$\theta$ in $[0,2\pi)$ with an interval of
$\frac{1}{18}\pi$, yielding a total of $36$ simulations.\KTC{Just to
verify, only one parameter, right? Please state.}\PYC{done.} We use $12$
of these $36$ simulations for training. The value of $\theta$ for these $12$ simulations
is evenly spaced with an interval of $\frac{1}{6}\pi$. The remaining $24$
simulations are used for testing. The goal of this training and testing split
is to gauge the ability of the proposed reduced-order model to
respond to pokes at arbitrary values of
$\theta$. Each simulation consists of $432$ time steps. Therefore, a total of
$15,588$ simulation snapshots, including the initial conditions, are used for
training and testing.

\begin{table}
  \centering
    \begin{tabular}{lrr}
      \toprule
      Scheme & Sample material & Sample material\\
      & points count: 50 & points count: 1,368 (all)\\
      \midrule
      Lagrangian quad + position-velocity proj & $1.47\%$ & $1.07\%$\\
      Lagrangian quad + velocity-only proj & $1.47\%$ & $1.04\%$\\
      Eulerian quad + position-velocity proj & $1.67\%$ & $1.41\%$\\
      Eulerian quad + velocity-only proj & $1.55\%$ & $1.37\%$\\
      \bottomrule
      \end{tabular}
      \caption{Poke-and-recover: errors of the reduced-order simulations on the testing dataset.} \label{tbl:poke-and-recover}
  \end{table}

  We first conduct offline training with $\lambda_F = 100, \lambda_v=0.01, \nred=6$ (cf. \Cref{sec:hyper-sum}) and then run online reduced-order simulations. \Cref{tbl:poke-and-recover} reports the testing errors of the reduced-order simulations using different quadrature and projection combinations with and without hyper-reduction, demonstrating the effectiveness of the reduced-order simulation in modeling the poke-and-recover problem.

\subsubsection{Reduced-space trajectory}\KTC{I don't see the value in  including this plot or section. It was
previously helpful for debugging, but this doesn't seem too relevant for the
paper.} \KTC{Peter: thoughts on removing this?}\PYC{5.3.1 is the explanation of why 5.3.2 works. I think it's important theoretical observation, for free manipulation, which is important for metaverse applications (that we all love). }
\begin{figure}[H]
  \centering
  \includegraphics[width=0.6\textwidth]{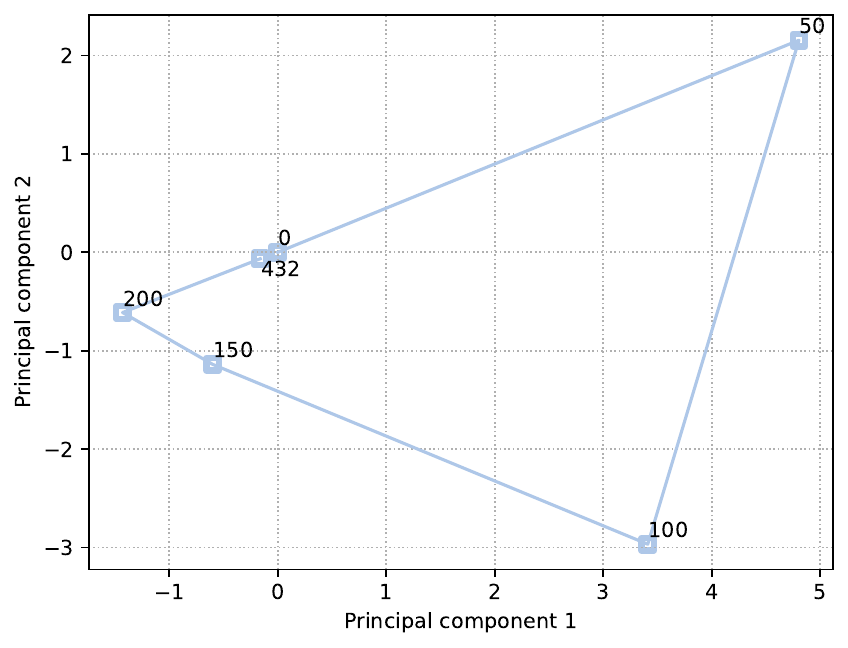}
  \caption{The temporal trajectory of the generalized coordinates $\genCoords$,
  visualized by its first two principal components. The time step number is
  annotated next to the trajectory. $\genCoords$ at time step 0 corresponds to the initial state (\Cref{img:poking-test} left); $\genCoords$ at time step 50 corresponds to the deformed state (\Cref{img:poking-test} middle); $\genCoords$ at time step 432 corresponds to the recovered state (\Cref{img:poking-test} right). Under the poking force, the generalized
  coordinates move away from their initial values and then return to their initial values due to elasticity.}
  \label{img:poke-and-recover:traj}
\end{figure}

\Cref{img:poke-and-recover:traj} plots the trajectory of the generalized coordinates $\genCoords$ of a reduced poke-and-recover simulation. Since $\genCoords$ is high-dimensional in general, we visualize $\genCoords$ by projecting it onto a 2D plane spanned by its first two principal components.

Under the influence of the poking force, the material takes up a deformed state in the full space $\overrightarrow{\vx}$; after the force is removed, the material then returns to its undeformed state. It is also desirable to maintain such a ``return'' property in the reduced space $\genCoords$, where $\genCoords$ returns to its initial value. In general, one state in the full space can correspond to multiple generalized coordinates, i.e., the approximate deformation map is not necessarily injective with respect to $\genCoords$. By using the encoder training scheme presented in \Cref{sec:encoder}, we can encourage injectivity and can indeed maintain the ``return'' property in the reduced space. As shown in \Cref{img:poke-and-recover:traj}, the generalized coordinates $\genCoords$ return to the origin, which maps to the undeformed state in the full space.

\subsubsection{Continual manipulation}
\begin{figure}[H]
  \centering
  \includegraphics[width=\textwidth]{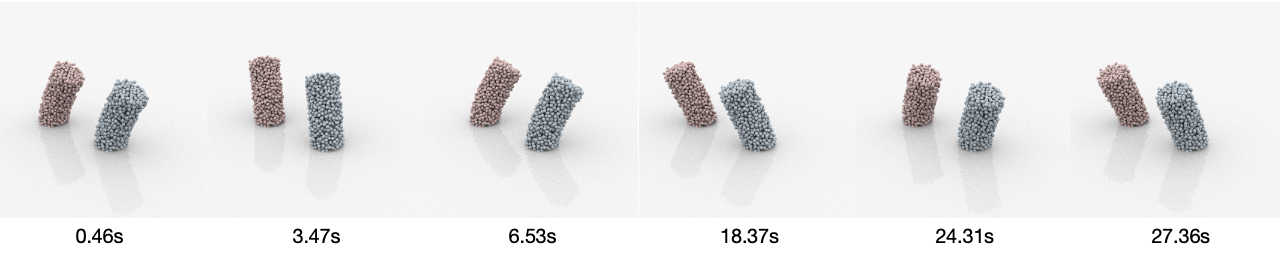}
  \caption{Repetitive poking. The reduced-order model supports continual manipulation of the object. We demonstrate several snapshots from a 30-second simulation sequence. The object on the left (red) is the ground truth; the object on the right (blue) is the reduced-order simulation. The reduced-order simulation agrees well with the full-order simulation.}
  \label{img:poke-and-recover:continual}
\end{figure}
\begin{figure}[H]
  \centering
  \includegraphics[width=\textwidth]{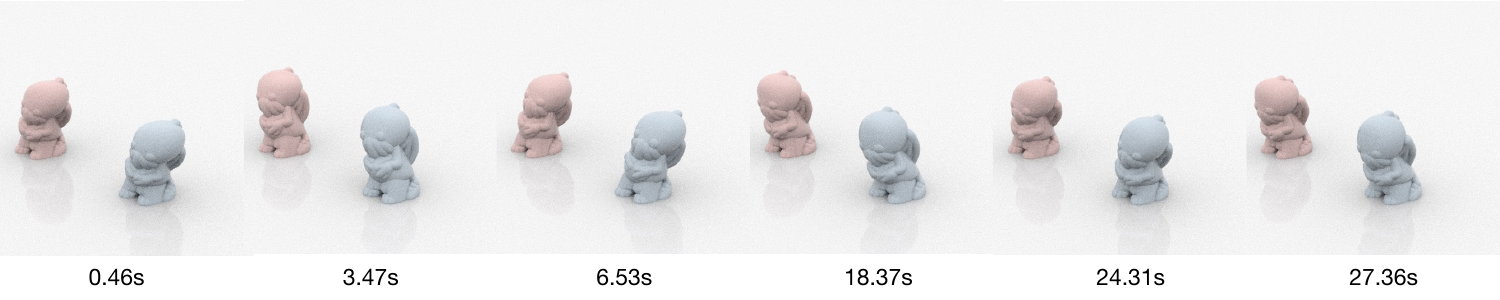}
  \caption{Repetitive poking of a more complex geometry \citep{wang2020hierarchical}. The object on the left (red) is the ground truth; the object on the right (blue) is the reduced-order simulation. \MMC{Can you crop the images so it's bigger - a lot of white space}\PYC{done.}}
  \label{img:poke-and-recover:continual-faceless}
\end{figure}
One advantage of maintaining the ``return'' property is that even though the training data consists of simulations poking only once, we can actually run reduced-order simulations that poke repetitively. After each poke-and-recover sequence, the generalized coordinates $\genCoords$ return to their starting values, ready to be poked again in an arbitrary direction.

Therefore, we run a reduced-order simulation consisting of ten consecutive poke-and-recover sequences, where the poking direction is chosen each time randomly (\Cref{img:poke-and-recover:continual}). This simulation has 4320 time steps or 30 s in total. \Cref{tbl:continual-poke-and-recover} reports its error in comparison with the full-order simulation. All quadrature and projection combinations produce good agreements with and without hyper-reduction. Note that no full-order simulation of 30 s is included in the training data. All training simulations are 3 s, demonstrating our framework's robust generalization capability. Our work also handles more complex geometries (\Cref{img:poke-and-recover:continual-faceless}). 

\begin{table}
  \centering
    \begin{tabular}{lrr}
      \toprule
      Scheme & Sample material & Sample material\\
      & points count: 50 & points count: 1,368 (all)\\
      \midrule
      Lagrangian quad + position-velocity proj & $1.72\%$ & $1.45\%$\\
      Lagrangian quad + velocity-only proj & $1.82\%$ & $1.41\%$\\
      Eulerian quad + position-velocity proj & $1.88\%$ & $1.81\%$\\
      Eulerian quad + velocity-only proj & $1.96\%$ & $1.74\%$\\
      \bottomrule
      \end{tabular}
      \caption{Continual poking and recovering: errors of the reduced-order simulations.} \label{tbl:continual-poke-and-recover}
  \end{table}

\subsection{Large-scale experiments}
\label{sec:application}
 
\begin{figure}
   \centering
   \includegraphics[width=0.8\linewidth]{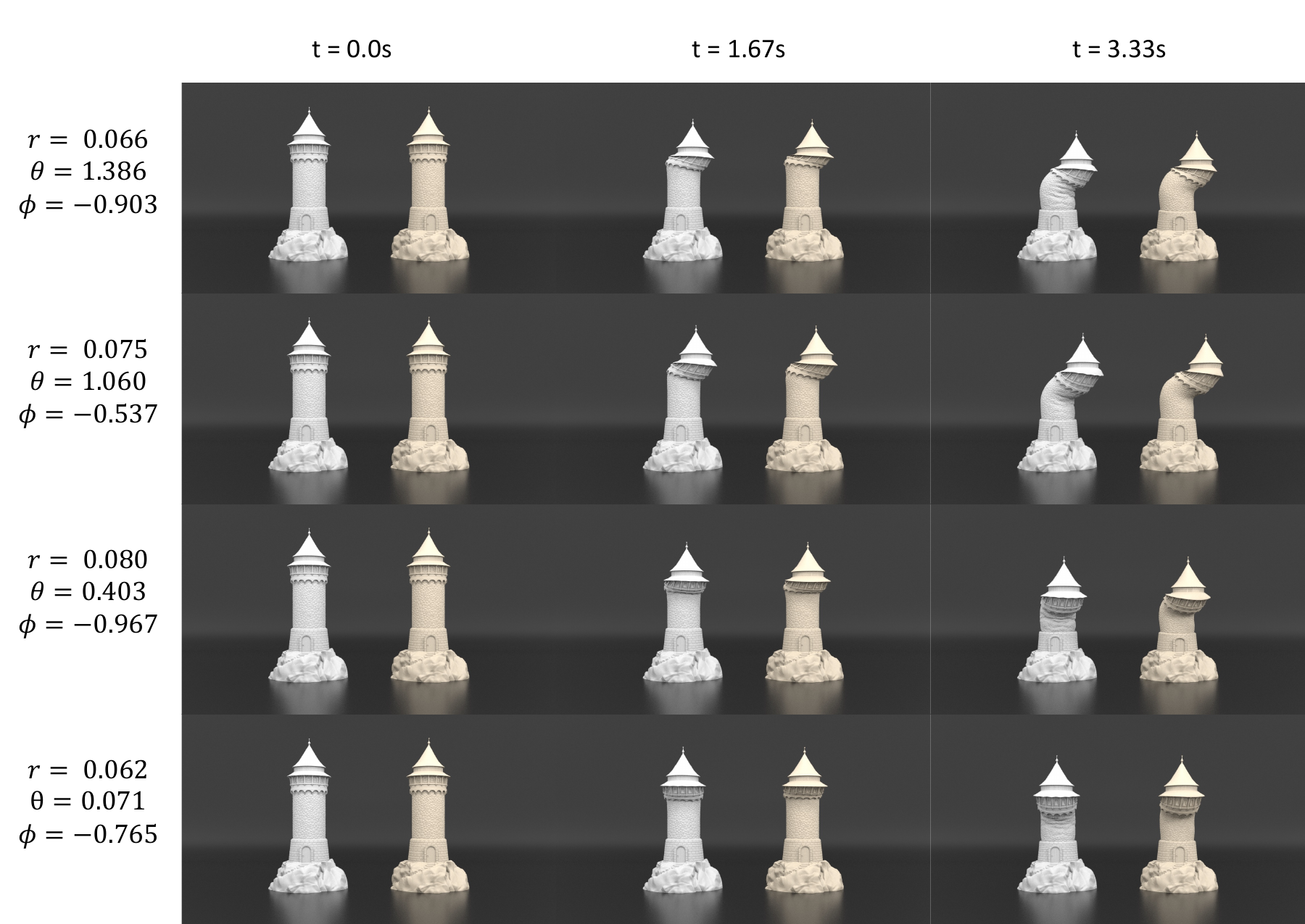}
   \caption{An object with complicated geometry undergoes elastic deformation (visualized with mesh, cf. \Cref{img:hyperreduction:visualize} for raw material point data). Each row records a different configuration in the testing dataset (i.e., unseen during training). Each column corresponds to a different time during the simulation. The leftmost column is the beginning of the simulation, while the rightmost column is the end of the simulation. In each snapshot, the white tower on the left is the full-order simulation, while the yellow tower on the right is the reduced-order simulation. The full-order model and the reduced-order model match overall in all configurations and at all times. However, the reduced model lacks secondary wrinkles that are present in the full-order model. More complex network architecture can be explored in order to capture these secondary features.}
   \label{img:large-deformation}
 \end{figure}

To demonstrate the efficiency of the reduced-order simulation in comparison to the full-order simulation, we conduct large-scale experiments with a complex tower geometry (\Cref{img:large-deformation}). 
The material and discretization parameters of the object are listed in \Cref{tbl:example_parameters}. \rev{\Cref{tbl:training-time} details the data generation time and training time.}

\begin{table}
  \centering
    \begin{tabular}{ll}
      \toprule
      Data generation time  & Training time\\
      \midrule
      $1140.8$ s & $35558.9$ s \\
      \bottomrule
      \end{tabular}
      \caption{\rev{Total offline cost: data generation and training. Data generation time includes generating 16 full-order simulations. Training time includes the time for training the manifold-parameterization function (\Cref{sec:manifold-construct}). Note that the timing data is purely informative and calculated post-hod whereas the main goal of our work is not to optimize training time.}} \label{tbl:training-time}
  \end{table}

Both the top and the bottom of the object experiences Dirichlet boundary
conditions. The top is kinematically moved under a fixed velocity while the
bottom is stationary.\KTC{`stays still' very colloquial}\PYC{fixed.} The fixed velocity is parameterized by a spherical coordinate, $\boldsymbol{v}=(r, \theta,
\phi)=(r\cos\phi\cos\theta,r\cos\phi\sin\theta,r\sin\phi)$, where $r \in
[0.6,0.8)$, $\theta \in [0,\frac{\pi}{2})$, and $\phi \in
[-\frac{2\pi}{3},-\frac{\pi}{3})$.

\KTC{Don't quite get the point of the first sentence here; can you clean up
this paragraph a bit? There's also passive voice that came back in.}\PYC{done.} We analyze the proposed approach in this parametrized setting ($\mub=(r, \theta,
\phi) \in \paramDomain\subseteq\RR{3}$). We generate training data by running 16 simulations with different $(r,\theta,\phi)$
triplets sampled using the Latin hypercube method \citep{stein1987large}.\KTC{Just to clarify, 3 parameters?}\PYC{done.} We further sample another $4$ simulations for testing purposes using the same approach.

\begin{figure}[H]
  \centering
  \includegraphics[width=\linewidth]{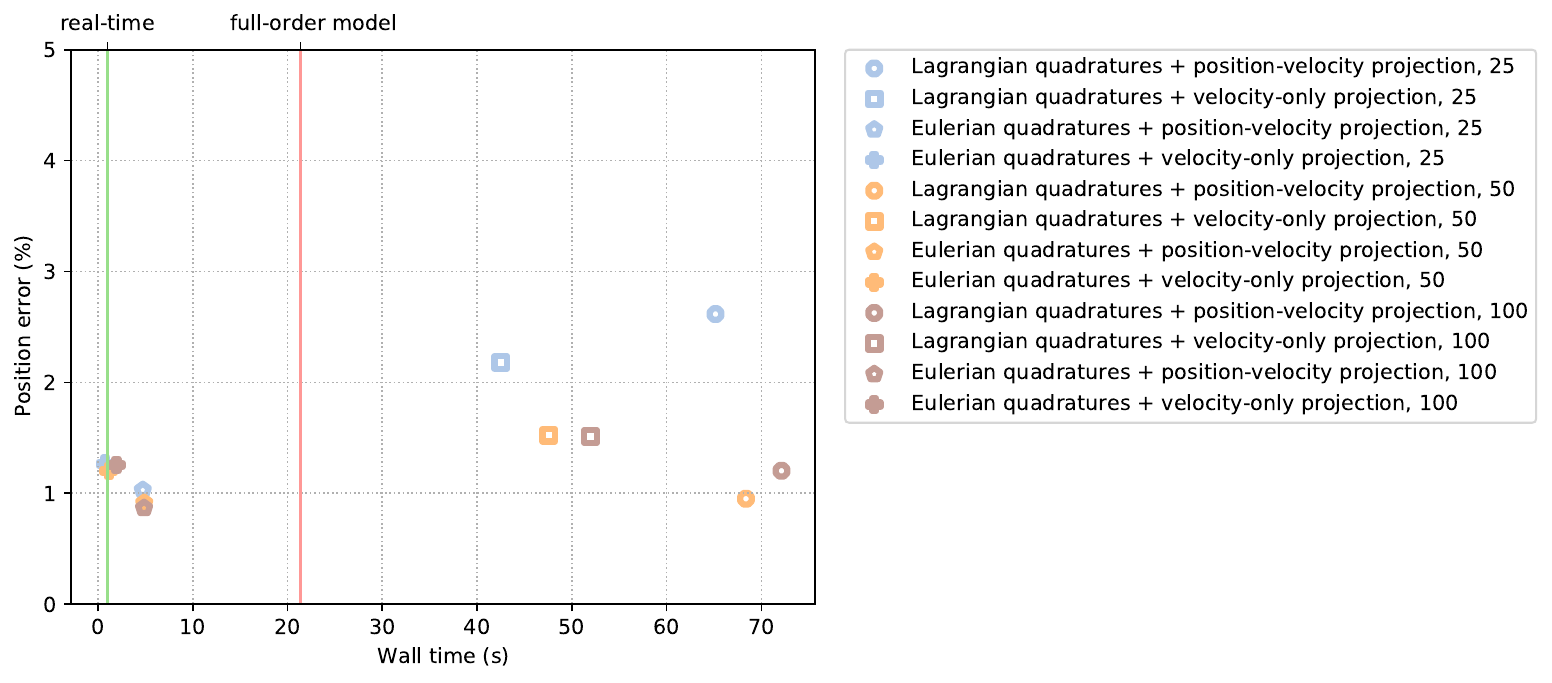}
  \caption{Position error vs. wall time. Each data point corresponds to a
  particular reduced-order setup (quadrature choices, projection types, and
  the number of sample material
  points). Wall time is the average computational
  cost for every physical second of simulation. A real-time simulation
  requires the wall clock time to be 1 (green line). The red line indicates
  the wall clock time of the original simulation. Setups using Eulerian
  quadratures and velocity-only projection with fewer than 50 sample material
  points (blue
  cross and orange cross) reach the real-time criteria and are over 20 times
  faster than the full-order model while maintaining accuracy. \KTC{I
  have a hard time visually distinguishing the blue and purple. Can you
  changer the colors to something more distinct?}\PYC{done.}}
  \label{img:large:pareto}
\end{figure}

An approximated deformation map network is trained with $\lambda_F = 100,
\lambda_v=0.01, \nred=6$ (cf. \Cref{sec:hyper-sum}). Afterward, we conduct
systematic tests over the different reduced-order approaches and different
numbers of sample material
points (\Cref{img:large:pareto}). All
setups using the Eulerian quadratures (pentagon and cross) offer a significant
speedup over the full-order model while maintaining accuracy. By contrast, due
to their need to track every material point, Lagrangian quadrature approaches
(circle and square) do not offer a reduction in computation complexity and do
not offer speedup over the full-order model. While velocity-only projection methods
(square and cross) generally have a slightly higher error than position-velocity
projection methods (circle and pentagon), they are also computationally
faster. Across all methods, the fewer sample material
points, the faster the
simulation and the higher the error. In particular, the reduced-order
simulations employing Eulerian quadratures and velocity-only projection with fewer
than 50 sample material
points (blue cross and orange cross) attain
\textit{real-time}
performance and are over 20 times faster than the baseline full-order
model. \Cref{img:large-deformation} displays all four testing simulations
using the Eulerian quadratures and velocity-only projection with 50 sample material
points
(orange cross). Visually, the reduced-order simulations agree well with the
full-order simulations while missing some secondary wrinkle deformations.
Further research can be conducted on increasing the complexity of the network
to capture these secondary deformations \citep{sitzmann2020implicit, takikawa2021neural}.

\begin{remark}
  Instead of computing the dynamics of over 3 million material points, we only
	need to calculate the dynamics of no more than 50 points (over 600,000 times
	reduction). However, the speedup number we observe is reduced to 20X. This
	discrepancy can be understood by the nonlocal nature of MPM, where a
	neighborhood of quadratures points is required for computing the dynamics of
	even just one material point. Consequently, in order to update the dynamics
	of 50 points, over 20,000 quadrature points are involved in the particle to grid transfer.\KTC{Any thoughts on future work to address this bottleneck,
	e.g., adaptive sparse-grid quadratures, distillation of the NN to run
	faster, optimizing the selection of sample material points, optimizing the
	implementation for parallelization? I think we haven't yet come close to the
	ceiling of speedup, and it'd be good to flesh out some ideas here of where
	to go next from a performance perspective.}\PYC{done. more in future work section.} To achieve the full wall-clock performance potential of the reduced-order model, further research should consider adaptive quadrature rules that lower the total number of quadrature points involved.  
\end{remark}

\begin{remark}
  We adopt a random sampling approach for choosing hyper-reduction sample material
  points (\Cref{sec:hyper-reduction}). While such a method is easy to implement, it does not guarantee optimality in terms of errors and computation costs. Future work should be conducted to select the optimal set of hyper-reduction points to minimize the position error and the computation cost.
\end{remark}

\subsubsection{Comparison}
\label{sec:application-comparison}
\KTC{Awesome you got this to work. Two things missing: (1) what is the
dimension of the linear ROM? \PYC{clarified, the same as nonlinear} (2) we should do a comparison as the reduced
dimension increases. Check out my paper with Kookjin Lee \cite{lee2020model}.
What's interesting is what happens as the linear ROM gets higher dimensional.
Any possibility to try $\nred\in\{1, 5, 7, 10\}$ or something that shows the
linear ROM is starting to improve? \PYC{I ran the experiment but no difference. all fail terribly. nonlinear activation is necessary for learning the \emph{continous} super nonlinear deformation map. this is very different from POD. see more on POD in the next comment. also in the extreme case where $\nred$ goes to infinity, we still cannot guarantee that we can approximate the continous function; BY contrast, if $\nred$ is the same as the degree of freedom of the discrete system, then we are guaraneed to approximate the discrete system with POD.}}
To compare the proposed framework with other model reduction methods, we first
notice that there is no prior work on model reduction of MPM using the
classical approach (\Cref{img:vs_classic}a). As discussed in \Cref{overview},
the classical approach (e.g., POD) is unsuitable for model reduction of MPM due to the
challenge of approximating the deformation gradients and achieving
hyper-reduction. Consequently, we construct a baseline model using our
approach (\Cref{img:vs_classic}b) with a linear manifold. Essentially, we
replace the nonlinear manifold-parameterization function with a linear one. In
practice, this amounts to replacing the multilayer neural-network-based
manifold-parameterization function (\Cref{img:network}) with a single linear
layer (without activation), wherein we train only the weights and biases of a
single layer; as such, it is similar to classical linear-subspace methods
such as POD.\KTC{Peter: please verify the last sentence here}\PYC{similar, yes. however, still very different tho. For example, you cannot use PCA/POD to contrsuct a \emph{continous} deformation map. i.e., you cannot construct a reduced space for the continuous map with SVD.} Note that the the nonlinear manifold and the linear manifold share the same reduced-dimension of $\nred=6$.

\begin{figure}[H]
  \centering
  \includegraphics[width=0.8\linewidth]{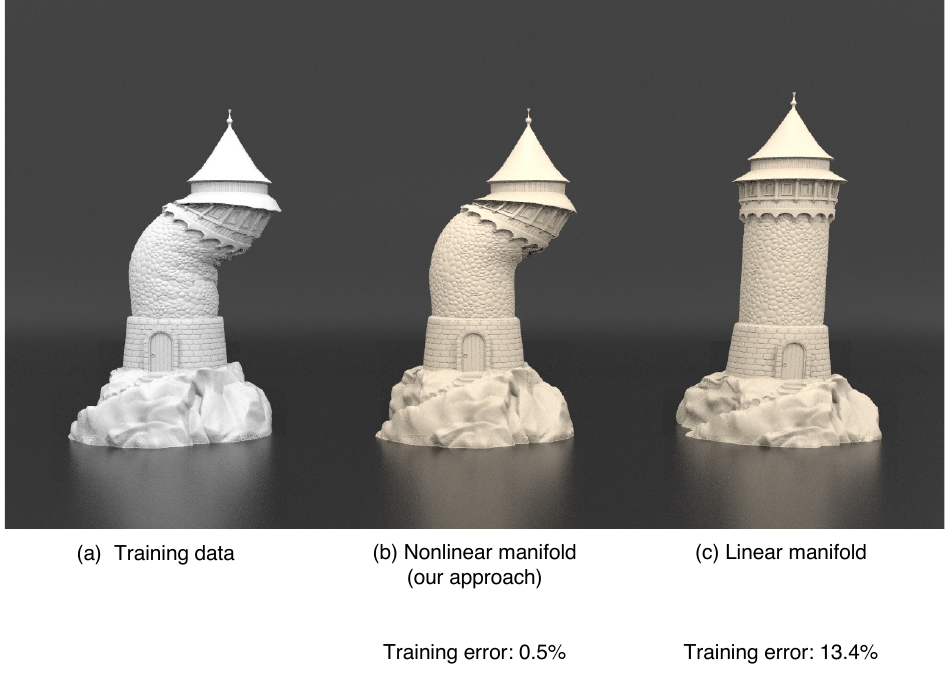}
  \caption{Nonlinear manifold vs. linear manifold. Replacing the nonlinear
	manifold-parameterization proposed in this work with a linear one leads to a
	significant performance decline. The linear manifold (c) struggles to
	approximate the highly nonlinear training data (a). By contrast, our
	proposed nonlinear manifold (b) accurately reconstructs the training data. The nonlinear manifold and the linear manifold both use a reduced-dimension of $\nred=6$.}
  \label{img:large:linear-manifold}
\end{figure}

As shown in \Cref{img:large:linear-manifold}, our nonlinear manifold significantly outperforms the linear one in terms of training accuracy, both quantitatively and visually. The linear approximation struggles to reconstruct the highly nonlinear deformation of the elastic object. We further attempted to deploy the trained linear manifold in the same setting as \Cref{img:large:pareto}. However, simulations using the linear manifold all went unstable in just a few time steps and were unable to complete the simulation. This is somewhat anticipated since the projection-based dynamics (\Cref{sec:dynamics}) further depend on the gradient information of the manifold-parameterization function. We thereby draw the conclusion that the robust, expressive nonlinear manifold (engineered via a deep neural network) proposed in this work is essential for building a manifold-parameterization function of the highly nonlinear deformation map.

\section{Conclusions and future work}
\label{sec:conclusions}
This work has presented---to our knowledge---the first projection-based
reduced-order model for the material point method. In contrast
with prior model reduction techniques that build a low-dimensional manifold of
the discretization of the ``deformation map'', we create a
discretization-agnostic, continuously-differentiable, low-dimensional manifold
of the ``deformation map'' itself based on implicit neural representations. We then utilize this low-dimensional
manifold to drive the MPM simulation via optimal-projection-based dynamics,
ensuring the simulated trajectory remains on the low-dimensional manifold
associated with the deformation-map approximation. We propose two different
quadrature approaches for computing full-space kinematics, and two different
projection approaches for computing reduced-space dynamics. Through the
introduction of hyper-reduction, we demonstrate that this
approach can drastically reduce the dimension of the MPM hyperelasticity
simulations and offers an order-of-magnitude wall-time speedup.

\MMC{ I would highlight that the contribution of this work is not training yet another neural network  but  is a framework for model reduction for MPM. Namely by building a $\hat d$-parameters family of functions expressive enough to capture the salient dynamics of our  dynamical system, through a three  step  approach, namely assembly (or something else for a name), full space dynamic update and projection, you enable the reduction of the dimensions of the system that along with quadrature reduction strategies allows to perform MPM simulations at a fraction (be quantitative here) of the FOM cost.}\PYC{Done, removed neural network, emphasized model reduction of deformation map.}

Moving forward, we envision 3 exciting directions to improve our work. (1) Supporting more continuum mechanics phenomena. We aim to extend our work to support other material behaviors, such as plasticity, fracture, contact, and collision. (2) Improving wall-time performance. This work focuses on the
\emph{spatial} model reduction of MPM. Future work should also consider a
reduction in the \emph{temporal} domain in order to take a larger time step
size. Since the stress evaluation is completed on the CPU while the neural
network evaluation is completed on the GPU, expensive CPU-GPU transfer is
conducted at each time step. Future work might investigate a full GPU
implementation to avoid the costly transfer. \rev{One may also consider expediting training time via more advanced data structures and optimization \citep{muller2022instant,liu2020neural,martel2021acorn,takikawa2021neural}. To improve training scalability, we can use hyperreduction samples as the input to the encoder instead of the full-order samples.} (3) Extending the use of implicit neural representations in model reduction for other types of systems and discretization methods. Even though the proposed model
reduction framework is designed for MPM, the proposed manifold
parameterization function is, in fact, discretization independent. Therefore,
we would like to go beyond MPM and explore its ability in model reduction of
other continuum mechanics discretizations, such as the finite element method
(FEM) and smoothed-particle hydrodynamics (SPH). For the same reason, we would
also like to explore the manifold parameterization function's ability to learn
from simulation data with adaptive refinement.
\KTC{I would do a more detailed and systematic breakdown of future work into
categories, e.g., (1) supporting more continuum mechanics phenomena, (2)
improving wall-time performance (see earlier comment, e.g., distillation\PYC{bottleneck is not network architecture}), (3)
extending the use of implicit neural representations in model reduction for
other types of systems and discretization methods.}\PYC{done.}

\section*{Acknowledgements} %
We thank Henrique Teles Maia for proofreading the manuscript. This work was supported in part by the National Science Foundation (Grants CBET-17-06689 and CHS-1717178) as well as SideFX.

\bibliographystyle{elsarticle-harv}
\bibliography{references}

\appendix
\section{Projection linearization}
\label{sec:linearization}
Substituting $\genCoords_{n+1} = \genCoords_n + \timestepn\genVels_{n+1}$ into
\Cref{eqn:nonlinear_projection:pos} yields
\begin{align}
    \label{eqn:pluginvhat}
    \genVels_{n+1} \in  \underset{\genVels\in\RR{\nred}}{\mathrm{arg\;min}}\ \sumParticlesSampleOnly\|
    \deformationMapApprox
    (\X^p;\genCoords_n+ \timestepn\genVels) - \xTnpTrial
    \|_2^2.
\end{align}
Using Taylor's theorem, we have
\begin{align*}
    \deformationMapApprox
    (\X^p;\genCoords_n+ \timestepn\genVels) - \xTnpTrial
    &\approx 
    \deformationMapApprox
    (\X^p;\genCoords_n) + \timestepn\frac{\partial
    \deformationMapApprox}{\partial
    \genCoords}
    (\X^p;\genCoords_n) {\genVels}- \xTnpTrial\\
    &=\deformationMapApprox
    (\X^p;\genCoords_n) + \timestepn\frac{\partial
    \deformationMapApprox}{\partial
    \genCoords}
    (\X^p;\genCoords_n) {\genVels}-({\xb^{p}_{n}} + \timestepn \velTnpTrial)\\
    &=\timestepn\frac{\partial
    \deformationMapApprox}{\partial
    \genCoords}
    (\X^p;\genCoords_n) {\genVels}-\timestepn \velTnpTrial\\
    &=\timestepn(\frac{\partial
    \deformationMapApprox}{\partial
    \genCoords}
    (\X^p;\genCoords_n) {\genVels}-\velTnpTrial)
\end{align*}
Under this linearization, \Cref{eqn:pluginvhat} becomes \Cref{eqn:linear_projection}. Consequently, the effectiveness of velocity-only projection depends on the accuracy of such a linearization.

\section{Training details}
\label{sec:training_details}
We implement the network in PyTorch \citep{paszke2019pytorch} and train the network with the ADAM optimizer with an adaptive learning rate, decreasing from $1e-3$ to $1e-6$. We initialize the neural network's weights using the Xavier initialization \citep{glorot2010understanding}. We conduct standard feature-scaling for the network's input and output to ease the training process. Min-max normalization is performed for the reference positions, while standardization is conducted for the network's output to have zero mean and unit variance.

\end{document}